%% file: main.tex
\documentclass{article}

\usepackage[final]{neurips_2023}

\usepackage[utf8]{inputenc} % allow utf-8 input
\usepackage[T1]{fontenc}    % use 8-bit T1 fonts
\usepackage{hyperref}       % hyperlinks
\usepackage{url}            % simple URL typesetting
\usepackage{booktabs}       % professional-quality tables
\usepackage{amsfonts}       % blackboard math symbols
\usepackage{nicefrac}       % compact symbols for 1/2, etc.
\usepackage{microtype}      % microtypography
\usepackage{xcolor}         % colors
\usepackage{wrapfig}
\usepackage{tikz}
\usepackage{graphicx}
\usepackage{float}
\usepackage{color}
\usepackage{colortbl}
\usepackage{bbm}
\usepackage{listings}
\usepackage{algorithm,algcompatible}
\usepackage{longtable}
\usepackage[caption=false]{subfig}
\usetikzlibrary{shapes}

\input{math_commands.tex}

\definecolor{citecolor}{HTML}{0071BC}
\definecolor{linkcolor}{HTML}{ED1C24}
\definecolor{LightGray}{gray}{0.8}
\usepackage{url}
\PassOptionsToPackage{table}{xcolor}
\def\eg{\emph{e.g.}}
\def\etc{\emph{etc}}

\title{Alternating Gradient Descent and Mixture-of-Experts for Integrated Multimodal Perception}

\author{Hassan Akbari\thanks{Equal contribution.} \quad Dan Kondratyuk\footnotemark[1] \quad Yin Cui \\
\vspace*{0.5em}
\textbf{Rachel Hornung} \quad \textbf{Huisheng Wang} \quad \textbf{Hartwig Adam} \\
Google Research\\
\texttt{\tt\small\{hassanak, dankondratyuk, yincui, rachelhornung, huishengw, hadam\}@google.com}
}

\begin{document}

\maketitle

\begin{abstract}
    We present Integrated Multimodal Perception (IMP), a simple and scalable multimodal multi-task training and modeling approach.
    IMP integrates multimodal inputs including image, video, text, and audio into a single Transformer encoder with minimal modality-specific components.
    IMP makes use of a novel design that combines Alternating Gradient Descent (AGD) and Mixture-of-Experts (MoE) for efficient model \& task scaling.
    We conduct extensive empirical studies and reveal the following key insights:
    1) performing gradient descent updates by alternating on diverse modalities, loss functions, and tasks, with varying input resolutions, efficiently improves the model.
    2) sparsification with MoE on a single modality-agnostic encoder substantially improves the performance, outperforming dense models that use modality-specific encoders or additional fusion layers and greatly mitigates the conflicts between modalities. 
    IMP achieves competitive performance on a wide range of downstream tasks including video classification, image classification, image-text, and video-text retrieval.
    Most notably, we train a sparse IMP-MoE-L focusing on video tasks that achieves new state-of-the-art in zero-shot video classification: 77.0\% on Kinetics-400, 76.8\% on Kinetics-600, and 68.3\% on Kinetics-700, improving the previous state-of-the-art by +5\%, +6.7\%, and +5.8\%, respectively, while using only 15\% of their total training computational cost.
\end{abstract}

%%%%%%%%%%%%%%%%%%%%%%%%%%%%%%%%%%%%%%%%%%%%%%%%%%%%%%%%%%%%
\section{Introduction}
\label{sec:introduction}

The human perception system is profoundly multimodal.
We perceive the world through the integration of a vast array of sensory systems across domains --- visual, auditory, olfactory, somatic, \etc.
Neurons for multimodal integration have been found in both multisensory convergence zones~\cite{calvert2001crossmodal} and unimodal regions~\cite{driver2008multisensory} in the human brain.
Studies in developmental psychology also suggest that interrelating simultaneous multimodal sensations is key for perceptutal learning~\cite{smith2005development}.
Inspired by these findings, we see an opportunity for combined multisensory learning in machine learning systems as well.

The rapid rise of large-scale multitask frameworks and models~\citep{raffel2020exploring, roberts2022scaling, radford2021learning, yu2022coca, wang2022image} provides foundations for integrating capabilities that unify many disparate tasks under one model.
However, given the vast quantity of independent variables involved in designing such a system, achieving an integrated multimodal machine learning model still remains an open research direction.
More specifically, designing a multi-task model that integrates many multimodal signals is challenging due to various reasons: i. Different modalities require structurally different I/O signatures to properly train. ii. When training across multiple datasets, some modalities or objectives may not exist or cannot be applied, depending on the input data and the task to perform. iii. The presence of multiple input modalities calls for careful considerations on the architectural design and allocation of parameters to certain modalities, often requiring extensive hyperparameter tuning to find the best use of computational resources.

Intuitively, as we scale a model, it becomes increasingly expensive to redesign the architecture or search for a better training objective.
The issue is exacerbated in multimodal multi-task modeling, where we need to consider the combination of input modalities or datasets, loss functions, and tasks at large scales.
Therefore, we would like to find a training approach that can be scaled incrementally: for any new task or objective, regardless of its input shape or output loss, we should be able to add it to the existing pretraining without compromising the previous tasks.

We navigate this problem by exploring ways in which we could train one multimodal model such that it (1) leverages as many existing datasets as possible, (2) can train on any combination of tasks or loss functions, and (3) does not slow down with the addition of any new dataset, task, or loss function.
By solving all of these points simultaneously, a multimodal model could scale with an increasingly diverse and rich set of training data without needing to redesign the training framework when new tasks get integrated.

We observe in our empirical results that the combination of diverse, heterogeneous tasks that have been previously established as strong objectives individually (\eg, supervised classification and self-supervised contrastive learning) across multiple modalities are not only complementary, but can offer better convergence than training on individual tasks. By implementing Alternating Gradient Descent (AGD) and Mixture-of-Experts (MoE) via recently developed JAX primitives, we enable our model to use a fraction of the computational cost and memory required by similar large-scale perception models~\citep{radford2021learning, jia2021scaling, yu2022coca}, despite the addition of multiple modalities which would normally require 2-8$\times$ compute at similar batch sizes.

Given this context, our contributions and findings are as follows:
\begin{enumerate}
    \item We define an integrated modality-agnostic encoder model, and leverage a strong combination of image-text contrastive, video-text contrastive, video-audio contrastive, and image/video/audio classification losses during pretraining to create an {\bf I}ntegrated {\bf M}ultimodal {\bf P}erception (IMP) model, as shown in Figure~\ref{figure:architecture}.
    \item Contrasting the conventional approach of summing the losses of multiple objectives, we show that alternating between objectives results in a design that allows seamless integration of virtually any number of tasks and datasets without significant memory overhead and results in better downstream evaluations.
    \item We show that optimization between multiple heterogeneous multimodal tasks is complementary and results in a higher quality model than trained on any individual task.
    \item To train on large batches of video and audio modalities without reducing training efficiency or loss of accuracy, we design a dynamic mixture of various resolutions, sequence lengths, and batch sizes throughout pretraining, and alternate training on all input variations.
    \item We enable our model with MoE, showing strong performance gains compared to a more conventional multi-tower contrastive model, even when appplying MoE to both towers.
    \item We scale our resulting MoE-IMP model to 2B sparse parameters with similar compute to ViT-L (300M parameters), resulting in state-of-the-art performance on several large-scale multimodal video understanding datasets.
\end{enumerate}

%%%%%%%%%%%%%%%%%%%%%%%%%%%%%%%%%%%%%%%%%%%%%%%%%%%%%%%%%%%%
\section{Related Work}
\label{sec:related_work}
The optimality of AGD optimization vs. averaging the losses (or gradients) has been explored in prior work~\citep{jain2017non, pascal2021improved}. Alternating multimodal multi-task training with AGD has been explored in PolyViT~\citep{likhosherstov2021polyvit}, which analyzes different methods to combining heterogeneous task in a single model. The work reports similar findings to our own, that combining objectives can be mutually beneficial and alternating between datasets weighted by the number of examples provides one of the best methods for optimization. Our work extends this to a much more generic setup supporting virtually any combination of modalities, tasks, resolutions, etc.

The use of sparse MoE for multimodal modeling can be seen in recent works like LIMoE~\citep{mustafa2022multimodal}, which uses a single MoE encoder for image-text tasks, and VL-MoE~\citep{shen2023scaling}, which uses modality-specific experts for image-text modeling.
Our work extends this concept further, introducing video and audio modalities with alternating training on multiple tasks and resolutions without requiring modality-specific experts.

Due to the inherent complexities of integrating modalities in one model, work has been done by simplifying the problem and focusing on a small set of universal objectives applicable to all modalities~\citep{yu2022coca,wang2022image}.
Alternatively, some works focused on applying padding or masking strategies to handle the incompatible I/O signatures from different tasks.
We found in either case, this severely limits the ability for a model to leverage pre-existing large-scale datasets or to scale to entirely new modalities. The historical reasons are that: (i) most existing models are designed for specific input modalities \eg, language~\citep{brown2020language,chowdhery2022palm}, vision~\citep{dosovitskiy2020image}, or audio~\citep{baevski2020wav2vec}; (ii) different modalities are typically delegated to separate network weights for the best performance~\citep{radford2021learning,jia2021scaling}; (iii) optimization difficulty with multiple modalities~\citep{wu2022scaling,chen2022pali}.
% \todo{add MoE comparison}

\section{Method}
\begin{figure}[t]
\centering
\subfloat[The IMP model architecture.]{\includegraphics[width=0.91\linewidth]{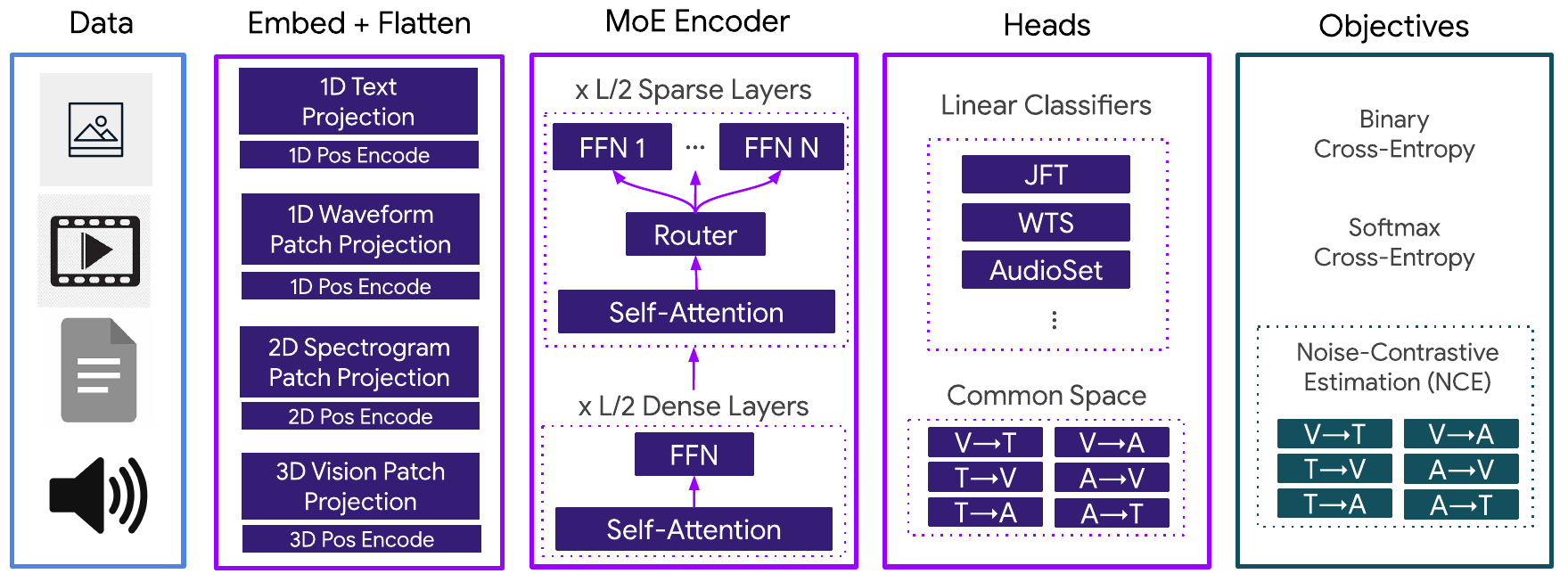}}
\hfill
\subfloat[The AGD-based multi-data multi-objective training overview.]{\includegraphics[width=0.92\linewidth]{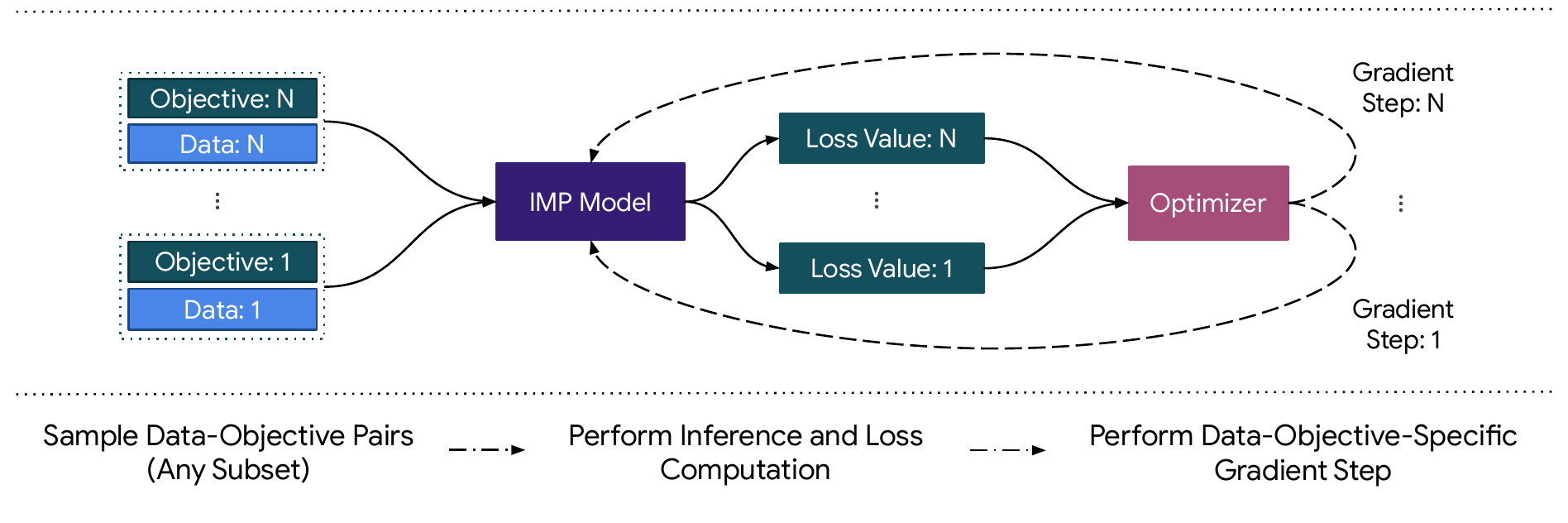}}
\caption{
        {\bf An overview of the IMP Training and Architecture.}
        A mixture of datasets with varying modalities, resolutions, and objectives are randomly sampled at each optimization step and fed into the model. The model is updated alternatingly given the data-objective pairs.
        We use {\tt jax.jit} to compile and cache computation graphs to keep each step efficient while also allowing I/O shapes to change for every optimization step without requiring any costly padding or masking strategies.
    }
\label{figure:architecture}
\vspace{-4mm}
\end{figure}

\subsection{Alternating Gradient Descent (AGD)}
\label{sec:agd}

One of the core pillars of our approach to multimodal understanding is {\bf task scalability}.
I.e., different combinations of data and loss objectives should be interchangeable throughout training, while the addition of any new data or objective should not cause memory or computation overhead.
We found that a common issue in training large-scale foundation models in a distributed setting is that input signatures and loss objectives need to be static to avoid major inefficiencies.
Accelerated graph compilation APIs allow for low-level graph optimizations that maximize hardware FLOPs utilization on distributed devices such as GPUs and TPUs, but come at a cost of requiring static I/O signatures.

\begin{algorithm}[t]
    \caption{Accelerated Multimodal AGD Algorithm}
    \begin{algorithmic}
        \STATE {\bf Input:} $M$ (model), $T$ (training steps), $X$ (dataset-objective pairs), $f$ (sampling function)
        \STATE {\bf Initialize:} Model state $S_t$
        \WHILE{$t \leq T$}
            \STATE Sample data-objective pair $(D_t, L_t) \in X$ according to sampling function $f(t, S_t)$
            \STATE Compute forward pass predictions $P_t = jit(M(D_t))$
            \STATE Compute backwards pass on loss $jit(L_t(P_t, D_t))$
            \STATE Update model state $S_t$
        \ENDWHILE
    \end{algorithmic}
    \label{algorithm:agd}
\end{algorithm}

One approach to handle the issue with the static input signature would be to use {\it mixed batching}, where all possible inputs are constructed, and inapplicable inputs for a given dataset are padded and outputs are masked accordingly at each training step.
However, this comes at a great efficiency cost, since the more tasks that are added the more time is spent computing on padded inputs. The issue with having multiple objective functions is usually resolved by {\it mixed mini-batching}, where a batch is divided to multiple mini-batches with their corresponding objective functions. The gradients for each mini-batch and objective function pair are accumulated across multiple mini-batches and the model weights are updated once using an aggregated gradient. However, this approach is also difficult to scale since the gradients across multiple mini-batches are accumulated in memory and per-task batch size naturally reduces as we add more tasks.

We propose a more generic solution based on AGD~\citep{jain2017non}, allowing any changes to the three elements of the optimization system: inputs, model, and objective.
AGD can be seen as a superset of the conventional SGD, where at each gradient step, a different objective may be optimized given different sets of model weights and/or input modalities. More specifically, any input modality with arbitrary shape could consume any subset of the model while focusing on minimizing any specific combination of the objective functions.
According to~\cite{jain2017non} it is proven that if each of such optimization steps are convex individually, an alternation between them leads to a guaranteed convergence; as opposed to mini-batching where gradient accumulation could result in sub-optimal results. From a technical point of view, this approach requires compiling multiple computation graphs, one for each unique optimization step.
To enable efficient execution of multiple computation graphs, JAX offers native Just-in-Time (JIT) compilation with the {\tt jax.jit} API\footnote{\scriptsize\url{https://jax.readthedocs.io/en/latest/jax-101/02-jitting.html}}, which compiles the graph-of-interest at runtime and compiles a new graph if a change in structure is seen in any of the next optimization steps.
Graphs themselves are cached by JAX in an in-memory lookup table so that tasks only need to be compiled once.
We empirically observe that graph memory consumption constitute a negligible portion of the entire training. In our experiments, we tested up to 20 unique task structures with no significant reduction in training speed, with compilation taking only $0.34\%$ of the total training time, up to the largest scales.

The main AGD algorithm for our multimodal framework is provided in Algorithm~\ref{algorithm:agd}. We note that we carefully design the loop to be as agnostic as possible with respect to the data-model-objective triplet. This allows for greater flexibility in defining logic inside the model to handle the processing of individual modalities and input shapes. The sampling function can also incorporate state from the optimization process itself. E.g., the loss value of a given step can be used as a reward signal to affect the sampling behavior in the next samplings~\citep{piergiovanni2023dynamic,mindermann2022prioritized}.
In our default setup, we sample each unique task on a given step from a (single trial) Multinomial distribution with probabilities directly proportional to the number of examples in each task. We defer more complicated or reward-based settings to future studies.

\subsubsection{AGD-Specific Efficiency Considerations}
We notice that in each forward call, certain model states such as activations are stored by default to be used later in the backward call for gradient calculation. This is an established compute optimization at the low-level graph compilation in XLA and similar APIs. Although this trick helps a lot with the training time, it significantly consumes memory and creates memory overhead if more than one graph is compiled and used during training.
To reduce memory overhead, we use JAX's native {\bf rematerialization} API, {\tt jax.checkpoint}\footnote{\scriptsize\url{https://jax.readthedocs.io/en/latest/\_autosummary/jax.checkpoint.html}} to save memory by not checkpointing any of the intermediate model states. In our experiments, we observe an average reduction of 70-80\% TPU HBM usage while resulting in only 18-20\% longer step times.

We also notice that large models with many different objectives may still incur long compilation times.
Therefore, we apply {\bf scan-over-layers} with {\tt jax.lax.scan}, a method which rolls all of the Transformer layers into a single layer that is called multiple times using different weights (instead of compiling the same function multiple times).
This alone results in 15-30x faster compilation time depending on the model length. We observe increasing relative time savings with larger model sizes.

Furthermore, we accomplish these across distributed accelerators through the {\tt jax.pjit} API\footnote{\scriptsize\url{https://jax.readthedocs.io/en/latest/notebooks/Distributed_arrays_and_automatic_parallelization.html}}, which distributes JIT compilation across multiple accelerators.

%%%%%%%%%%%%%%%%%%%%%%%%%%%%%%%%%%%%%%%%%%%%%%%%%%%%%%%%%%%%
\subsection{Objectives}
\label{sec:objectives}

Our goal in designing the IMP model is to reuse objectives that have been shown to be robust for learning each modality. Hence, we choose the two most established supervised and unsupervised learning objectives: i. Supervised Classification using Softmax/Binary Cross-Entropy(SCE/BCE), ii. Cross-Modal Noise-Contrastive Estimiation (NCE). Unless otherwise specified, we do not sum any of the above losses or accumulate gradients as would be done traditionally. Instead we apply backprop on each objective individually with AGD.

\subsection{Architecture}
\label{sec:architecture}
Figure~\ref{figure:architecture} shows a high-level overview of the architecture of IMP, which consists of three main modules: i. The {\bf Embedder}, which accepts specific modalities and embeds them in a shared modality-agnostic space; ii. The {\bf MoE Encoder}, which computes semantic contextual embeddings from the embedded tokens; iii. The {\bf Heads}, which produce all the final predictions from the Encoder by re-projecting its embeddings back into a modality-specific space. We briefly explain each module here and provide more details in Appendix.

One design decision important to multimodal modeling is how to allocate parameters to each modality.
As seen in works like BASIC~\citep{pham2021combined}, an asymmetric modality-specific design can be more optimal than using a similar-sized model for each modality.
However, this comes at the cost of requiring additional hyperparameter tuning to find the optimal parameterization.
As we show later in the next section, we observe that through the use of model sparsification with MoE, a unified encoder design coupled with certain modality-specific pre- and post-encoder layers is more optimal than a traditional multi-encoder setup as seen in CLIP model variants.

We follow VATT~\citep{akbari2021vatt}, AudioMAE~\citep{huang2022masked}, and T5~\citep{raffel2020exploring} to extract the vision, audio, and text embeddings, respectively. We add learnable 3D/2D/1D positional encodings to the embeddings and project them to a space with the same dimensionality as the model's. We pass these embedded inputs regardless of modality as-is through the shared encoder, which is a standard Transformer architecture equipped with Mixture-of-Experts FFN layers. We follow V-MoE~\citep{riquelme2021scaling} and LIMoE~\citep{mustafa2022multimodal} for expert allocation. This can be seen as an inductive bias, allowing each expert to be allocated to multiple modalities if the optimization benefits. One immediate benefit is that the addition of new modalities for fine-tuning does not need any specific changes to the encoder, unlike modality-specific experts which require additional modifications and input handling~\citep{wang2022image, shen2023scaling}.

We apply modality-specific heads on the encoder representations to produce the final outputs for loss and prediction calculations.
For classification objectives, we apply a dataset-specific linear classifier to the average-pooled outputs.
For noise-contrastive estimation (NCE), we closely follow the CLIP architecture, applying separate feedforward heads for each modality-to-common-space projection.

\subsection{Multi-Resolution Training}

One major issue when training Transformers on video data is that computation and memory efficiency are usually bottlenecked due to Transformer's quadratic complexity as a function of the input length.
To counteract this, we propose to adjust batch size or resolution to compensate the additional temporal tokens, hence achieving a similar total number of input tokens compared to a single-frame still image.

We first fix a set tokens per batch $T = B \times T_F \times T_H \times T_W$, which is factorized by the batch size $B$, frame tokens $T_F$, height tokens $T_H$, and width tokens $T_W$ representing each patchified video.
We observe that we can further factorize each batch by trading off different dimensions such that the total number of input tokens per step are roughly equal so that peak memory usage is preserved.
For example, we can halve the spatial resolution while quadrupling the number of frames. This can increase convergence especially at the start of training, and provide a more memory efficient encoding of each objective. Furthermore, we leverage DropToken~\citep{akbari2021vatt} as an additional method to reduce tokens per batch by randomly dropping a fixed ratio of tokens per example. We find that for $T_F$ temporal frame tokens, we can randomly drop a ratio of $1 - \frac{1}{T_F}$ tokens per example to match the same tokens per batch as images. For certain objectives we find that a different mix of trade-offs is more optimal. For example, contrastive objectives favor large batch sizes, so we reduce the resolution or apply DropToken to be more memory efficient. On the other hand, classification objectives do not need as large batch sizes for optimal convergence, hence we reduce the batch size while increasing the spatiotemporal tokens.

%%%%%%%%%%%%%%%%%%%%%%%%%%%%%%%%%%%%%%%%%%%%%%%%%%%%%%%%%%%%
\section{Experiments and Results}
\label{sec:experiments}
\subsection{Training Setup}
\label{sec:training-setup}

\paragraph{Datasets.}
Our datasets consist of a diverse set of learnable signals across multiple modalities. We use WebLI~\citep{chen2022pali}, LAION-400M~\citep{schuhmann2021laion}, WIT~\citep{srinivasan2021wit}, CC12M~\citep{changpinyo2021conceptual}, and VCC~\citep{nagrani2022learning} for vision-text contrastive learning; JFT-3B~\citep{zhai2022scaling}, I21K~\citep{ridnik2021imagenet}, and WTS-70M~\citep{stroud2020learning} for both supervised classification and label-based vision-text contrastive estimation (similar to BASIC~\citep{pham2021combined}); HT100M~\citep{miech2019howto100m} and AudioSet~\citep{gemmeke2017audio} for vision-audio-text triplet contrastive loss (similar to VATT~\citep{akbari2021vatt}).

We use a proportionally weighted sampling algorithm, executing each task in succession. To ensure that datasets are evenly sampled, we weight each task by the number of examples, normalized to a probability distribution. For each dataset variant with different resolution sizes, we apply the same weight.
For a fair evaluation on downstream tasks, we filter all near-domain examples from our pretraining datasets (about 5M examples total).

\paragraph{Multi-Resolution Strategy.}
In our experiments, we always configure the input parameters so that the number of frame tokens are always equal to 4. This will result in the base tokens per video batch being exactly 4x of image's.
For video datasets, we construct three variants and uniformly sample from each variant during training: i. Reduce the resolution by half in each dimension, ii. Reduce the batch size by 4x, iii. Apply DropToken $d = 1 - \frac{1}{T_F} = 0.75$. For image datasets, we also apply a similar strategy but for the purpose of high-resolution learning. In addition to the base resolution, we have two extra variants: i. Reduce the batch size by 4x and double each spatial dimension, ii. Apply DropToken $d = 1 - \frac{1}{4} = 0.75$.

\paragraph{Training Parameters.}
For our final experiments, we train with a patch size of 4x16x16 on base input resolutions of 16x256x256 and 4x256x256 on video and image modalities respectively, resulting in a total of 1024 and 256 patches per sample.
The text inputs in ImageNet21K and JFT are truncated to 16 tokens to improve step efficiency with no loss of information, while keeping the text length of the rest of the datasets to a maximum of 256 tokens.
We use a base batch size of 65536 and train using the Adam optimizer, a peak learning rate of 1e-3 with a cosine schedule, and apply no weight decay.
For MoE parameters, we apply experts-choose routing with a top-$c$ capacity factor of 1.0 and do not apply any jittering to the routing or other auxiliary losses. Training results in roughly 16B examples seen, or about 5T tokens. Taken together, these datasets represent about 11B unique image-text and video-audio-text examples.

During inference, we evaluate on the largest available resolution that the model was trained on, i.e., 16x512x512, and use a total of 8 clips per video at approximately 12.5 fps on all evaluated datasets.

\subsection{Main Results}
\vspace{-1mm}
We scale up and tune IMP for best performance on video datasets and evaluate it on a variety of downstream tasks and datasets to understand how it generalizes to other modalities. Table~\ref{table:main-results-zero-shot} shows the zero-shot classification capabilities of the model on several image, video, and audio datasets.

\begin{table}[t]
    \begin{center}
    \resizebox{\columnwidth}{!}{
    \begin{tabular}{@{}lrrrrrrrrrrr@{}}
    \toprule
        \sc Model & \sc PPT & \sc TPU-days & \sc IN1k & \sc C100 & \sc K400 & \sc K600 & \sc K700 & \sc UCF101 & \sc HMDB51 & \sc ESC-50\\
    \midrule
        CLIP~\citep{radford2021learning} & 400M & - & 76.2 & - & - & - & - & - & - & -\\
        CoCa-B~\citep{yu2022coca} & 380M & 1.8k & 82.6 & - & - & - & - & - & - & -\\
        X-CLIP~\citep{ni2022expanding} & 400M & - & - & - & 65.2 & - & - & 72.0 & - & -\\
        BIKE~\citep{wu2022bidirectional} & 230M & - & - & - & - & 68.5 & - & 80.8 & 52.8 & -\\
        Text4Vis~\citep{wu2022transferring} & 230M & - & - & - & 68.9 & - & - & 85.8 & - & -\\
        AudioCLIP~\citep{gowda2021smart} & 430M & - & - & - & - & - & - & - & - & \bf 69.4\\
    \midrule
        {\bf IMP-B}     & 86M   & 120 & 80.5 & 82.4 & 63.6 & 62.1 & 49.9 & 64.2 & 39.7 & 32.8\\
        {\bf IMP-MoE-B} & 90M  & 150 & 83.2 & 84.9 & 68.2 & 65.7 & 52.1 & 88.7 & 46.6 & 47.8\\
        {\bf IMP-MoE-L} & 350M  & 1.5k & \bf 83.9 & \bf 87.0 & \bf 77.0 & \bf 76.8 & \bf 68.3 & \bf 91.5 & \bf 59.1 & 65.1\\
    \midrule
        \textbf{Large-scale models} \\
        LIMoE~\citep{mustafa2022multimodal} & 680M & - & 84.1 & - & - & - & - & - & - & -\\
        LiT ViT-g~\citep{chen2022pali} & 2B & - & 84.5 & 83.6 & - & - & - & - & - & -\\
        CoCa~\citep{yu2022coca} & 2B & 10k & 86.3 & - & - & - & - & - & - & -\\
        VideoCoCa~\citep{yan2022video} & 2B & 10.5k & - & - & 72.0 & 70.1 & 62.5 & 86.6 & 58.6 & -\\
    \bottomrule
    \end{tabular}
    }
    \end{center}
    \caption{
        {\bf Zero-Shot Classification} (top-1) results on image, video, and audio datasets.
        IMP achieves a new state-of-the-art on zero-shot video action recognition by a wide margin with significantly low training cost. Considering the total number of Parameters Per Token (PPT), IMP also significantly outperforms comparable models on zero-shot image classification.
    }
    \label{table:main-results-zero-shot}
\vspace{-4mm}
\end{table}

We note that IMP significantly outperforms previous state-of-the-art regardless of the model size and sets new record on Kinetics~\citep{kay2017kinetics, carreira2018short, carreira2019short}, UCF101~\citep{soomro2012ucf101}, and HMDB-51~\citep{kuehne2011hmdb} top-1 accuracy.
Compared to the previous state-of-the-art, VideoCoCa~\citep{yan2022video}, we train IMP-MoE-L on 256 TPU v4 chips for 6 days, representing only 15\% of the total training cost of VideoCoCa. Considering the total number of parameters per token (PPT), IMP also outperforms the previous comparable state-of-the-art model, CoCa, on ImageNet~\citep{ILSVRC15} and CIFAR-100 with a relatively large margin. However, we observe that the model falls behind the state-of-the-art in zero-shot audio classification on ESC-50~\citep{piczak2015esc}. This might be explained by the fact that the total training examples for audio modality are almost negligible compared to image and video. Hence, the model has a very strong performance on video and image. We argue that this could be resolved by simply introducing more samples and a more balanced train scheduling method, which we differ to future studies.

\subsection{Ablation}
\label{sec:ablation}
\vspace{-1mm}
In this section, we hightlight experimentation with some key results which motivate the chosen set of features for our final IMP model. We refer the reader to Appendix for more ablation on several other aspects of the model. The experiments in this section use IMP-S or IMP-B trained for 250k steps with a base batch size of 8192. We set a fixed video/image resolution of 16x224x224/4x224x224 using a patch size of 4x16x16. Unless otherwise specified, we do not apply multi-scale resolution.
\vspace{-1mm}
\paragraph{Combined objectives are mutually beneficial.}
We train the model on ImageNet21k with two objectives: Noise-Contrastive Estimation (NCE) and Softmax Cross-Entropy (SCE) and explore the following: i. train on the objectives separately, ii. combine the objectives by summing them, and iii. alternating (AGD) between the objectives on each step.
In the case of alternating, for a fair comparison so that training time is equivalent, we fix the same number of steps (250k) so that each objective only optimizes 50\% of the total steps (125k). We evaluate on ImageNet1k and CIFAR-100 by image-to-text retrieval and linear probing on the frozen model's features and report the results in Table~\ref{table:I21K-objectives}.
\begin{table}[t]
    \begin{center}
    \resizebox{0.65\columnwidth}{!}{%
    \begin{tabular}{@{}lrrrr@{}}
    \toprule
        \sc & \multicolumn{2}{c}{\sc ImageNet1K} & \multicolumn{2}{c}{\sc CIFAR-100} \\
        \sc Objective & \sc Linear & \sc I $\rightarrow$ T & \sc Linear & \sc I $\rightarrow$ T \\
    \midrule
        NCE                & 39.8 & 46.7 & 84.0 & 49.8 \\
        Softmax            & 41.1 & - & 82.6 & - \\
        NCE + Softmax, Sum & 47.6 & 46.7 & 82.4 & 51.6 \\
        NCE + Softmax, Alternating & \bf 49.9 & \bf 48.0 & \bf 84.1 & \bf 52.4 \\
    \bottomrule
    \end{tabular}
    }
    \end{center}
    \caption{
        {\bf Combining multiple objectives during pre-training.}.
        Alternating between both objectives offer the best performance, despite training on the same number of total steps.
    }
    \label{table:I21K-objectives}
\vspace{-4mm}
\end{table}
It is not a surprise that classification objective benefits fine-tuning evals the most, while contrastive objective benefits open vocabulary classification. However, we observe that combining both objectives is better than optimizing on them individually. And alternating between the objectives is better than non-AGD objective mixing. These results are similar to the findings of PolyViT~\citep{likhosherstov2021polyvit}, which report optimal performance on alternating the objectives, weighted by the size of each dataset.
{\it This motivates us to fix one objective per training step and alternate optimization between them.}

\begin{figure}[t]
\centering
\includegraphics[width=0.85\linewidth]{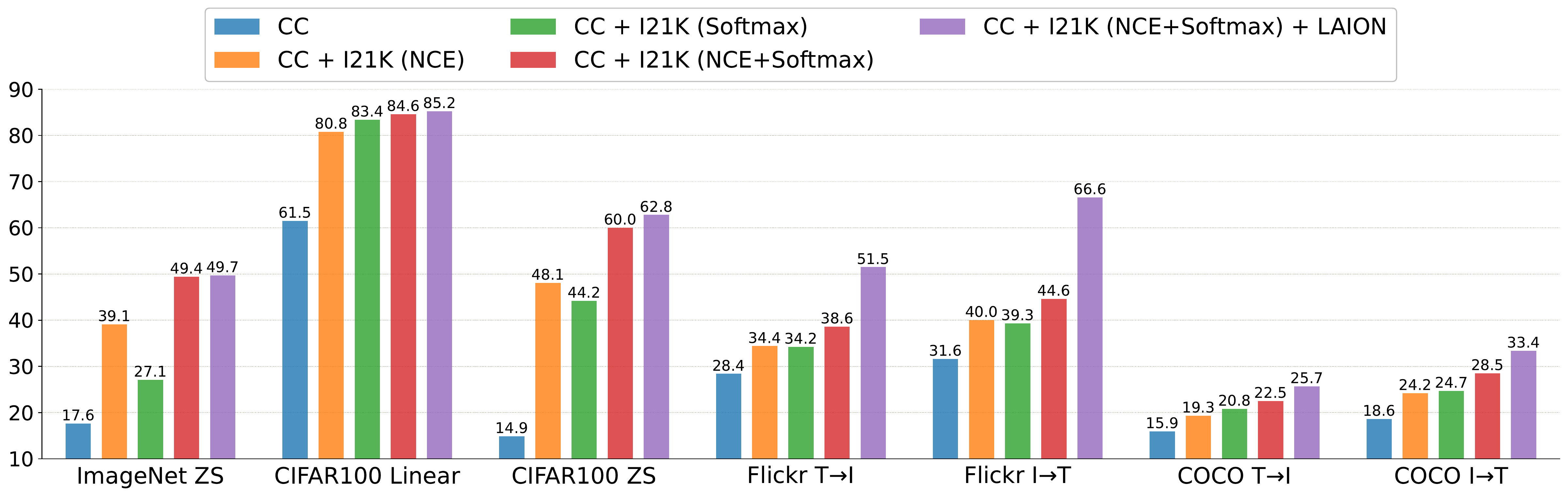}
\caption{
    {\bf Combining multiple datasets and objectives using AGD}.
    We integrate CC, I21k, and LAION with NCE and SCE using AGD and observe consistent improvement in downstream results. We also observe that NCE and SCE are mutually beneficial. Further optimality is provided by adding larger and more diverse datasets like LAION.
}
\label{figure:datasets}
\end{figure}

\paragraph{Multi-task multi-dataset AGD is also mutually beneficial.}
In Figure~\ref{figure:datasets}, we compare the result of adding additional datasets to the pretraining mixture.
We additionally compare results across Flickr30k~\citep{young2014image} and COCO~\citep{lin2014microsoft} datasets.
We start with CC12M dataset and gradually add new datasets and objectives. Most notably, we compare the addition of I21K dataset, showing complementary improvement when combining NCE and SCE objectives. Similar to I21K isolated experiments, adding SCE benefits the entire pretraining mixture. While SCE benefits zero-shot results, NCE benefits linear probing results too.
Certain dataset combinations (CC+I21K, CC+LAION) cause instability at the beginning of training.
Adding a classification objective has a stabilizing effect, significantly reducing the chance of slow convergence. Optimizing on LAION directly is difficult, but benefits training a lot more when mixed in with other datasets.
{\it This motivates us to further integrate a larger set of diverse datasets and objectives.}

\begin{figure}[!t]
\centering
\includegraphics[width=0.85\linewidth]{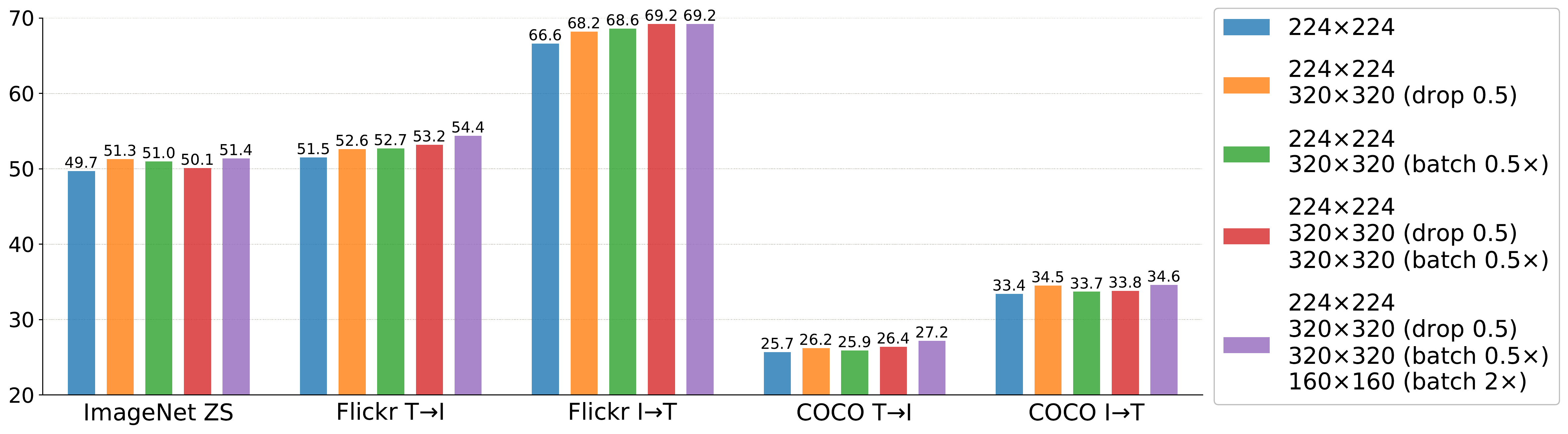}
\caption{
    {\bf Combining multiple input sizes using AGD}. We sample each variant with equal probability. Downstream results improve as we add more variants to the training mixture.
}
\label{figure:resolution}
\vspace{-4mm}
\end{figure}

\paragraph{Multi-scale resolution provides universal improvement.}
Figure~\ref{figure:resolution} shows a comparison of using different combinations of resolution, batch size, and DropToken as input. In all settings, we fix the total tokens per batch, and we ensure that all training runs use the same number of total steps. We see that certain types of datasets respond well to DropToken while others may not.
CC with double the batch size and DropToken 0.5 improves zero-shot image classification.
Droptoken + 320x320 image on I21K SCE pretrain is better for linear probing and image-text retrieval.
Adding multiple versions of smaller batch size + higher res, DropToken + higher res, larger batch size + lower res, can significantly improve downstream results.
{\it We find that dynamic mixtures of resolution, batch size, and DropToken are always helpful.}

\begin{figure}[t]
\centering
\includegraphics[width=0.75\linewidth]{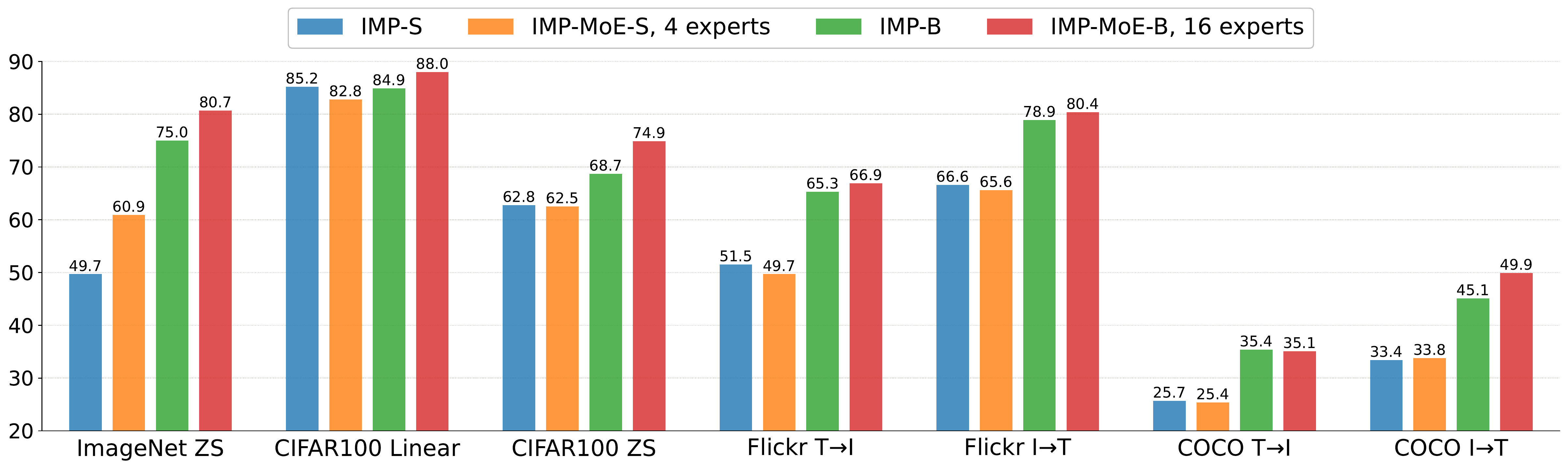}
\caption{
    {\bf IMP with Mixture-of-Experts}.
    Results show that using a modest 4 experts increases the model's accuracy substantially on ImageNet zero-shot evaluation.
    When we scale up the experts to 16, we see a consistent and significant improvement across all downstream evaluations.
}
\label{figure:moe}

\end{figure}

\begin{figure}[t]
\centering
\includegraphics[width=0.98\linewidth]{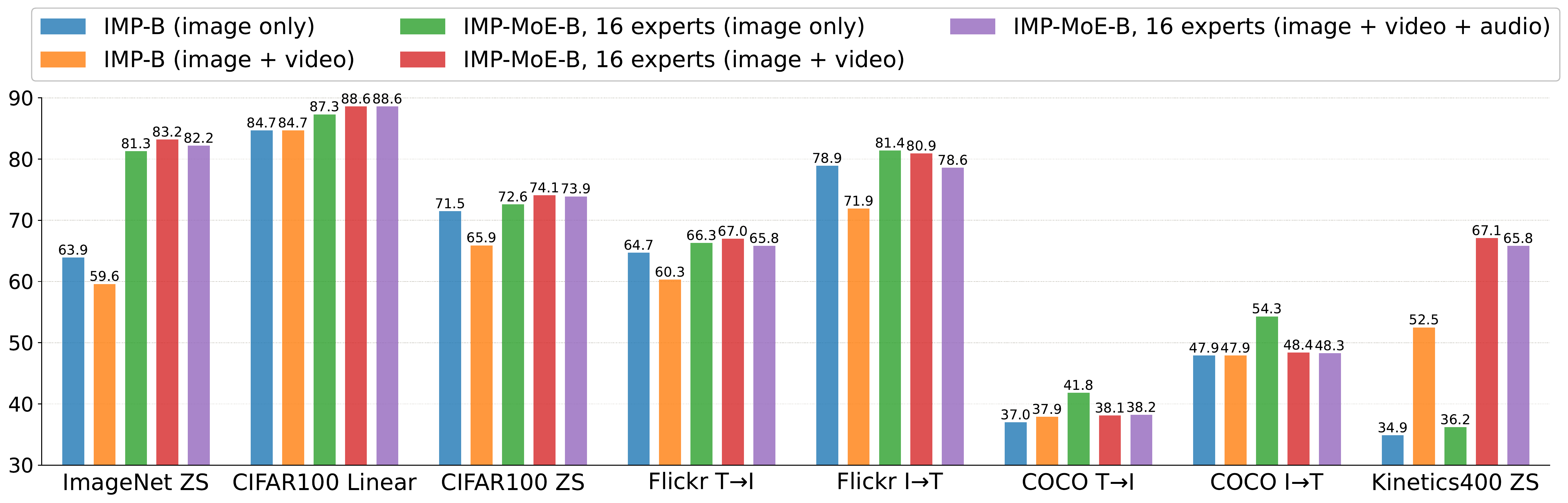}
\caption{
    {\bf Improved multimodal perception using MoE}.
    Results show significant improvement on diverse multimodal perception when we use the MoE variant of IMP. The addition of audio reduces accuracy on image and video metrics across the board, but is much less prominent when using MoE.
}
\label{figure:moe-multimodal}
\end{figure}

\paragraph{MoE provides universal improvement across modalities, and resolves the single-tower encoder parameter bottleneck.}
The main challenge in designing a unified encoder tower as we have described is that parameters must be split between multiple modalities, thus harming accuracy.
Compared to a two-tower contrastive model, the encoder of a unified image-text model contains half the parameters, while keeping training efficiency the same.
One direction we explore is whether a large increase in parameters from MoE is sufficient to resolve parameter bottlenecks.
In Figure~\ref{figure:moe}, we observe that simply replacing a dense model with an equivalent MoE model with just 4 experts, we can provide a large gain in accuracy, especially for zero-shot metrics.
This provides a promising indication that MoEs can be used to bridge the multimodal gap. We observe that with the addition of MoE, we can significantly close the gap between multiple modalities as seen in Figure~\ref{figure:moe-multimodal}.
Since experts are free to choose which tokens are allocated to different experts, we observe strong alignment between experts and modalities.

\begin{figure}[t!]
\centering
\includegraphics[width=0.75\linewidth]{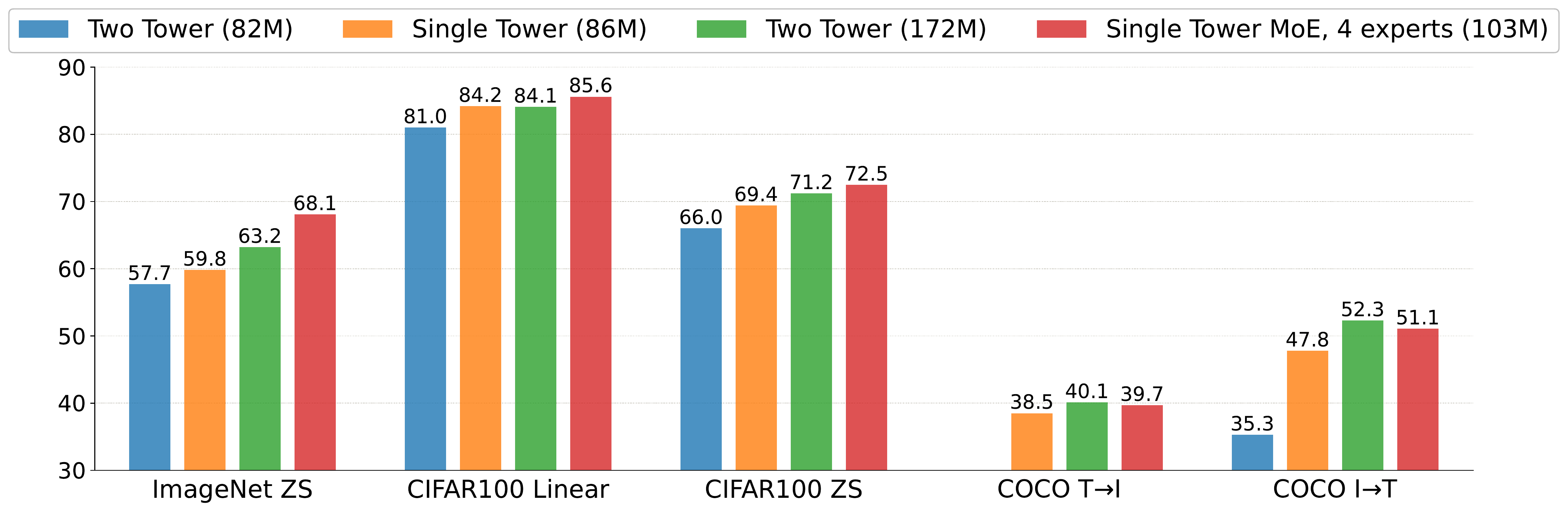}
\caption{
    {\bf Comparison of single-tower vs. multi-tower designs on IMP-B}.
    A single-tower MoE model is both parameter efficient and compute efficient compared to multi-tower dense variants.
}
\label{figure:towers}
\vspace{-0.8em}
\end{figure}
\vspace{-0.5em}

\paragraph{Single-tower MoE outperforms multi-tower dense variants.}
We find that out of all the variants we tested, a unified MoE encoder provided the most parameter and compute efficient design, while significantly outperforming a multi-encoder modality-specific model in downstream results, as seen in Figure~\ref{figure:towers}.
When comparing two-tower models, we can either split the parameters to be roughly equal in size to a single tower, or duplicate the towers to double the parameter count while providing equivalent computation. We observe that multi-tower dense models are more compute efficient than single-tower dense models, but less parameter efficient.
However, a single-tower MoE model is both more compute and parameter efficient than all variants, showing improved generalization and using fewer parameters with the same compute budget as a multi-tower dense model.
These results show universal superior parameter and compute efficiency and higher accuracy by using just 4 experts. This observation suggests that our method can be used for integrated multimodal multi-task modeling without worrying about the complications of modality-specific design choices or downstream performance degradation as observed in previous modality-agnostic designs \citep{akbari2021vatt}.

%%%%%%%%%%%%%%%%%%%%%%%%%%%%%%%%%%%%%%%%%%%%%%%%%%%%%%%%%%%%
\section{Conclusion}
\label{sec:conclusion}
In this paper we presented an integrated training and modeling approach for multimodal perception using AGD and MoE. We observed that AGD enables task scalability and multi-resolution training, which improves the training convergence and the model's generalization capabilities. On the other hand, we found that MoE can play a very important role in integrating multiple modalities into one unified model. Given these findings, we scaled the model with hyperparamters tuned specifically for video understanding and achieved state-of-the-art performance in zero-shot video action recognition with a significant margin. Furthermore, we observed that the model also generalizes on other modalities and achieves competitive downstream results. 
In a nutshell, IMP opens a door to data (\eg, modality, resolution, \etc.) and task scalability --- two important directions that have been neglected in many multimodal understanding works due to the inherent technical limitations. Due to the vast range of elements involved in this system, we defer multiple directions to be explored in future work: 1. generative objectives and model architectures, 2. causal MoE for generation, 3. sophisticated methods for data-objective sampling, 4. more downstream evaluations.

%%%%%%%%%%%%%%%%%%%%%%%%%%%%%%%%%%%%%%%%%%%%%%%%%%%%%%%%%%%%
\begin{ack}
We would like to thank Joan Puigcerver, Carlos Riquelme, and Basil Mustafa for their advice on MoE implementation and analysis; Anselm Levskaya for his help with advanced core JAX and Flax implementation; the T5X team for their support for scalable model partitioning, and Erica Moreira and Victor Gomez for their help with resource allocation.
\end{ack}

%%%%%%%%%%%%%%%%%%%%%%%%%%%%%%%%%%%%%%%%%%%%%%%%%%%%%%%%%%%%
\setcitestyle{numbers}
\bibliographystyle{plainnat}
\bibliography{main}

%%%%%%%%%%%%%%%%%%%%%%%%%%%%%%%%%%%%%%%%%%%%%%%%%%%%%%%%%%%%

\newpage
\input{appendix.tex}

\end{document}

%% file: math_commands.tex
\usepackage{amsmath,amsfonts,bm}

\def\eqref#1{equation~\ref{#1}}
\def\1{\bm{1}}

\DeclareMathAlphabet{\mathsfit}{\encodingdefault}{\sfdefault}{m}{sl}
\SetMathAlphabet{\mathsfit}{bold}{\encodingdefault}{\sfdefault}{bx}{n}

%% file: appendix.tex
\appendix

%%%%%%%%%%%%%%%%%%%%%%%%%%%%%%%%%%%%%%%%%%%%%%%%%%%%%%%%%%%%
\section{Architecture}
\subsection{Embeddings}
For vision modalities, we use the VATT~\citep{akbari2021vatt} scheme to patchify each 3D video tensor.
We define a video tensor of size $F \times H
\times W$ with $F$ frames and $H \times W$ resolution using a patch size of $f \times h \times w$, producing $\frac{F}{f}\times \frac{H}{h}\times \frac{W}{w} \times 3$ voxels that are flattened into a single sequence.
This is followed by linearly projecting the sequence into the model's hidden size.
To allow for more robust generalization, we treat images as a special case of a video, assuming sequences are of shape $f\times H\times W$ and tiling frames $f$ times to fit in a single patch.
For our base model, we use a patch kernel size of 4x16x16, up to 16 frames, and resolutions up to 512x512.
Following VATT, to allow the model to adapt to different resolution scales, we apply learnable positional encodings to each patch position, which consist of the sum of 3 embedding tables along each separate axis: one for the temporal dimension, and two for the spatial dimensions of the video patches.

For text, we apply T5 encoding~\citep{raffel2020exploring} using the default English vocabulary with 32k SentencePiece tokens, which are embedded into the same hidden space size as image patch embeddings.

For audio, we apply both waveform and audio spectrogram as input.
For spectrogram, following AudioMAE~\citep{huang2022masked}, after downsampling the audio waveform to 16000 kHz, we extract Mel-spectrograms with a duration of 8 seconds, producing 128 feature vectors with 128 dimensions each.
We apply a patch kernel size of 16x16 to produce 64 total patches as input.
For waveform, we use a kernel size of 256 samples and embed up to 256 tokens.

We use a separate learned positional embeddings for the linear sequence of text tokens and audio patches similar to the video encoding scheme above. To be able to handle different numbers of patches across different dimensions, the positional encoding of vision modalities needs to be handled with special care.
Unlike the 1-dimensional sequences of text and audio waveform which can be truncated to a given length, the presence of 2D spatial dimensions mean that images with double the patches along a dimension should be subdivided into quadrants so that adjacent positions are close to each other in the embedding space.
We accomplish this using a {\it dilated positional encoding}. For a given dimension a spatial positional encoding of $B$ buckets, if we encode a resolution with $P$ patches, we dilate the positional encoding with a stride of $\frac{B}{P}$. We treat spectrograms as 2D images and apply the same dilated encoding logic to them accordingly.
In the case of the temporal dimension in video, we treat it the same as a 1-dimensional truncation independent of the spatial dimensions, which is applied in the same way for text.

\subsection{MoE Encoder}
For all MoE encoders, we use expert-choice routing~\citep{zhou2022mixture}, which provides a strong baseline for all of the four modalities.
Using expert-choice (top-$c$) routing, we observe much higher accuracy compared to the standard tokens-choose (top-$k$) routing.
This is because experts-choose routing guarantees even load balancing, which we find to be an important factor for using an encoder shared across modalities.
We find that only applying MoE to the last 50\% of layers provided similar accuracy to applying them for all layers, therefore we use this setting for all MoE model variants.

We observe that contrastive optimization with MoE produces unstable output, often when introducing noisy text labels.
This instability results in a loss divergence roughly within the first 30k-80k training steps.
Similar to ViT-22B~\citep{dehghani2023scaling}, we find that applying a layer normalization after the key and query matrices (QK LayerNorm) in the self-attention layers removes all such divergence issues in our training, hence we use this trick in all of our model variants.

\subsection{Heads}
We apply global average pooling operation across the entire output sequence of the encoder, and use the resulting vector as the global features for classification and noise-contrastive estimation objectives.
For classification objectives, we apply a dataset-specific linear classifier to the average-pooled outputs.
For noise-contrastive estimation (NCE), we closely follow the CLIP architecture, applying separate feedforward heads for each modality-to-common-space projection. Each feedforward head consists of a two-layer linear projection with GeLU activation in between. The projection dimension size is the same as the model's hidden size.

\subsection{Model Sizes}
Table~\ref{table:architecture} provides a description of model sizes.
We provide results for three main variants, IMP-S, IMP-B, and IMP-L corresponding to encoder sizes of ViT-S, ViT-B, and ViT-L respectively~\citep{zhai2022scaling}.
We also provide three additional sparse MoE sub-variants, which are indicated as IMP-MoE.

\begin{table}[t]
    \begin{center}
    \resizebox{0.9\columnwidth}{!}{%
    \begin{tabular}{@{}lrrrrrr@{}}
    \toprule
        \sc Model & \sc Params (Dense) & \sc Params (Sparse) & \sc \# Experts & \sc \# Layers & \sc Hidden Size & \sc FFN Size \\
    \midrule
    \bf IMP-S       & 21M  & 40M  & 4   & 12 & 384  & 1536 \\
    \bf IMP-B       & 86M  & 400M & 16  & 12 & 768  & 3072 \\
    \bf IMP-L       & 300M & 2B   & 16  & 24 & 1024 & 4096 \\
    % IMP-8B      & 1B  & 8B    & 16  & 32 & 1536 & 8192 \\
    \bottomrule
    \end{tabular}
    }
    \end{center}
    \caption{
        {\bf Comparison of IMP Architectures}.
        We provide parameters for dense and sparse MoE variants.
        Note that we only apply MoE to the last half of the layers in the encoder.
    }
    \label{table:architecture}
\end{table}

\section{Training Setup}
\subsection{Datasets}
For large-scale pretraining, we use the following datasets:
\begin{enumerate}
    \item WebLI~\citep{chen2022pali} consisting of 4B English-only image-text pairs. We use this dataset for image-text contrastive loss.
    \item JFT-3B~\citep{zhai2022scaling}, which contains a large collection of multi-class labels per image. We follow BASIC~\citep{pham2021combined} for encoding multiclass indices as text and use the dataset for image-text contrastive loss as well as supervised classification loss.
    \item LAION-400M~\citep{schuhmann2021laion}, a public dataset of 400M image-text pairs for image-text contrastive loss.
    \item Wikipedia Image Text (WIT)~\citep{srinivasan2021wit} with 37M image-text pairs sourced from Wikipedia for image-text contrastive loss.
    \item Conceptual Captions (CC12M) ~\citep{changpinyo2021conceptual} consisting of 12 M image-caption pairs, used for image-text contrastive loss.
    \item ImageNet21K (I21K)~\citep{ridnik2021imagenet} with 11M labeled images for image-text contrastive loss and supervised classification loss.
    \item VideoCC (VCC)~\citep{nagrani2022learning}, a video dataset with a variant expanded to 1B English video-text pairs for video-audio-text triplet contrastive loss.
    \item HowTo100M (HT100M)~\citep{miech2019howto100m} consisting of $\sim$100M video-audio-ASR triplets, used for video-audio-text triplet contrastive loss.
    \item Weak Text Supervision (WTS-70M)~\citep{stroud2020learning}, a dataset of 70M video clips obtained based on 700 action classes. We use this variant for video-text contrastive loss as well as supervised classification loss similar to JFT-3B and IN21k.
    \item AudioSet~\citep{gemmeke2017audio} for video-audio-text triplet contrastive loss.
\end{enumerate}

%%%%%%%%%%%%%%%%%%%%%%%%%%%%%%%%%%%%%%%%%%%%%%%%%%%%%%%%%%%%
\section{Additional Ablation}
\begin{figure}[t]
\centering
\includegraphics[width=0.5\linewidth]{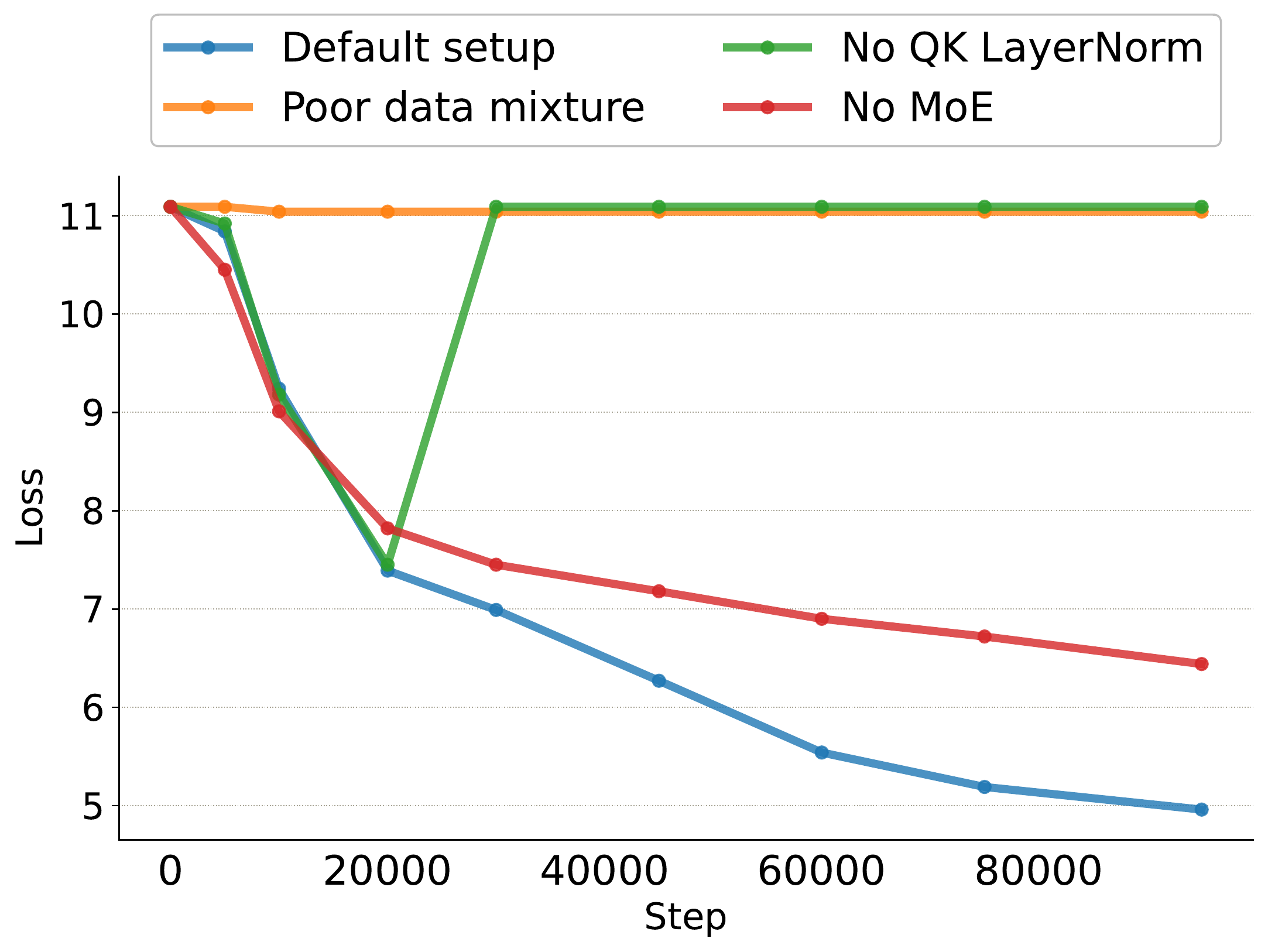}
\caption{
    {\bf Plot of loss with different training setups} in the first 100k steps of IMP-MoE-B.
    We observe that our MoE model with QK LayerNorm and using a diverse mixture of datasets reduces instability and produces the best loss convergence.
    A poor data mixture (e.g., CC + LAION) can cause a loss plateau, while adding QK LayerNorm in self-attention is important for avoiding loss divergence early in training.
}
\label{figure:moe-loss}
\end{figure}

\paragraph{Instabilities of contrastive loss on MoEs can be reduced with diverse data mixtures and QK LayerNorm.}
We observe that a combination of MoE training with contrastive losses can lead to divergence, as seen in Figure~\ref{figure:moe-loss}.
As seen in the figure (see also Table~\ref{table:datasets}), the addition of multiple datasets, even under the same objective, can be detrimental to the optimization process.
At the beginning of contrastive training on CC and LAION datasets, we observe a {\it loss plateau}, where the loss remains relatively constant from the start of training, and the model fails to start converging for a long period of time.
On the other hand, if we apply the same dataset but add a softmax objective from the ImageNet21K dataset, we no longer observe a loss plateau, as softmax tends to be more stable for optimization processes than contrastive losses.
% And if we integrate both softmax and NCE objectives on ImageNet21K, the plateau is also missing.
This highlights the importance of selecting the right dataset mixture, especially at the start of training where the inherent nature of the random parameters can make it difficult for some task gradients to solidify a good direction in the optimization process.

We also observe certain divergences which occur early in training, and we found the magnitude of gradient updates can get large in the attention inputs.
This can cause training to completely destabilize, and enter a similar loss plateau.
Therefore, we apply QK LayerNorm, which we observe to have prevented such divergence across all of our experiments.

% \begin{figure}[htbp]
% \centering
% \includegraphics[width=0.6\linewidth]{figures/bar_chart_towers_moe_i1k.pdf}
% \caption{
%     {\bf Comparison of single tower MoE designs} on IMP-S, comparing experts-choose (top-$c$) and tokens-choose (top-$k$) approaches.
%     Results indicate that single-tower top-c 
% }
% \label{figure:moe-towers}
% \end{figure}

\begin{figure}[t]
\centering
\includegraphics[width=0.95\linewidth]{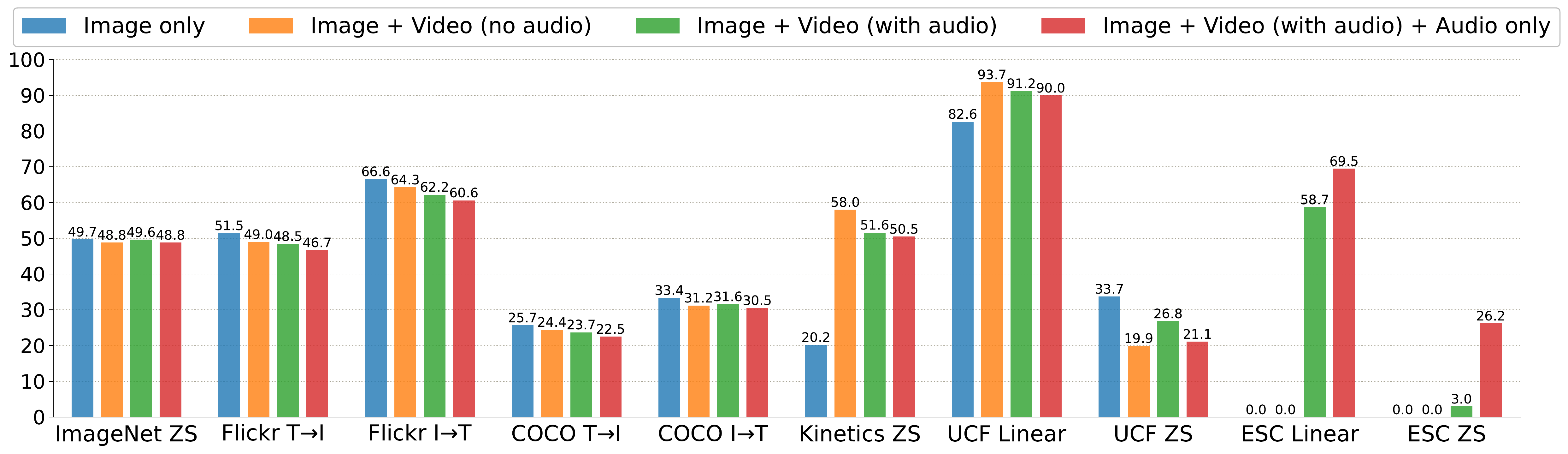}
\caption{
    {\bf Comparison of the addition of video and audio datasets on IMP-S}.
    The addition of video data (i.e., WTS) substantially improves video accuracy (Kinetics400, UCF) at the cost of slightly reducing the model's accuracy on image tasks (ImageNet, Flickr30k, COCO).
    The introduction of audio in the contrastive loss (i.e., AudioSet) also reduces both image and video accuracy slightly, but enables fine-tuning on audio data (ESC).
    Finally, the introduction of a dedicated audio class contrastive objective hurts image and video accuracy the most, but enables zero-shot audio classification.
}
\label{figure:video}
\end{figure}

\paragraph{Adding more modalities hurt single tower (dense) encoder accuracy.}
We compare the addition of more modalities via video datasets in Figure~\ref{figure:video}.
Adding a video dataset (i.e., WTS) to pretraining boosts Kinetics classification significantly, allowing the model to more easily discriminate between action classes.
However, the addition of video data may harm image classification performance slightly, especially when parameters are constrained.
Likewise, the addition of audio data with video has a slight negative impact, and the addition of dedicated audio classification dataset (i.e., AudioSet) has an even larger negative impact.
This may be due to the additional audio-text contrastive signal which requires the network to allocate dedicated processing for audio-to-text understanding.
{\it Therefore, a standard single tower encoder is not sufficient for optimal multimodal learning due to parameter bottlenecks.}

\paragraph{Experts-choose routing is crucial for strong single tower performance.}
We test the effect of experts-choose routing vs. tokens-choose in Table~\ref{table:moe-towers}.
Similar to findings in VL-MoE~\citep{shen2023scaling}, we observe that separating experts by modality in the case of tokens-choose routing is useful for improving accuracy.
However, when we switch to experts-choose routing, we find that performance increases further, and a multi-tower model is similar enough in accuracy that separating them per modality is no longer necessary.
This allows for a much simpler model design, and we can fine-tune new modalities in the encoder without any additional setup or new experts required.

\begin{table}[t]
    \begin{center}
    % \resizebox{\columnwidth}{!}{%
    \begin{tabular}{@{}ccccc@{}}
    \toprule
        \sc \# Towers & \sc \# Experts & \sc Router & \sc Params & \sc ImageNet ZS \\
    \midrule
        1 & 1 (dense) & N/A & 86M & 59.8 \\
        1 & 4 & tokens-choose & 103M & 62.5 \\
        2 & 4 & tokens-choose & 206M & 65.7 \\
        1 & 4 & experts-choose & \bf 103M & \bf 68.1 \\
        2 & 4 & experts-choose & 206M & 68.4 \\
    \bottomrule
    \end{tabular}
    % }
    \end{center}
    \caption{
        {\bf Comparison of single tower MoE designs on IMP-B}, comparing experts-choose and tokens-choose approaches on ImageNet class retrieval.
        The most accurate and parameter efficient configuration is a single-tower experts-choose model.
        For tokens choose, we use a maximum capacity factor of 1.05 as in V-MoE so that training times are roughly equivalent.
    }
    \label{table:moe-towers}
\end{table}

\paragraph{Inserting diverse prompts during training helps improve zero-shot classification.}
In Table~\ref{table:prompts}, we evaluate different prompt settings during pretraining.
Contrasting with prior works which only apply prompts during evaluation, we observe strong gains when randomizing the prompts during training.
One interesting exception is in CIFAR-100, which benefits from training and testing on no prompts at all.
We observe the trend that simpler prompts are more useful for datasets with a smaller number of classes.
We typically care more about results of large-scale datasets, so we apply prompt diversification by default.

\begin{table}[htbp]
    \begin{center}
    \resizebox{\columnwidth}{!}{%
    \begin{tabular}{@{}lrrrrrrrrr@{}}
    \toprule
        \sc Configuration & \sc IN Linear & \sc IN ZS & \sc C100 Linear & \sc C100 ZS & \sc F30k T$\rightarrow$I & \sc F30k I$\rightarrow$T& \sc UCF Linear \\
    \midrule
        No Prompt & 46.6 & 51.4 & \bf 83.9 & \bf 67.5 & 50.3 & \bf 66.8 & 83.0 \\
        CLIP Prompt & \bf 47.7 & 55.8 & 83.7 & 57.8 & 49.7 & 66.5 & 81.9 \\
        IMP Prompt & \bf 47.7 & \bf 56.5 & 83.8 & 60.5 & \bf 51.2 & 66.3 & \bf 84.1 \\
    \bottomrule
    \end{tabular}
    }
    \end{center}
    \caption{
        {\bf Prompt comparison of IMP}.
        We compare three settings of train-time prompts with increasing diversity.
        Results show that randomized prompts in training tend to significantly improve metrics on classification tasks.
    }
    \label{table:prompts}
\end{table}

\paragraph{Stability of optimization.}
In general, we observe the following situations where optimization can become more unstable: (1) Increase in dataset diversity; (2) Increase in model size; (3) Increase in batch size.
In Table~\ref{table:datasets}, we show that training on NCE alone can cause instability issues during training, especially for noisy text datasets.
But with the addition of more clean data sources and softmax objectives, we can greatly reduce the instability.

% \todo{jax code of MAX training loop}

\begin{table}
    \begin{center}
    \resizebox{\columnwidth}{!}{%
    \begin{tabular}{@{}lrrrrrrrrr@{}}
    \toprule
        \sc Datasets & \sc IN Linear & \sc IN ZS & \sc C100 Linear & \sc C100 ZS & \sc F30k T$\rightarrow$I & \sc F30k I$\rightarrow$T & \sc COCO T$\rightarrow$I & \sc COCO I$\rightarrow$T  & \sc UCF Linear \\
    \midrule
        CC                              & 15.2 & 17.6 & 61.5 & 14.9 & 28.4 & 31.6 & 15.9 & 18.6 & 73.3 \\
        CC + LAION                     & Diverged & - & - & - & - & - & - & - & - \\
        CC + I21K (NCE only)            & 33.5 & 39.1 & 80.8 & 48.1 & 34.4 & 40.0 & 19.3 & 24.2 & 78.3 \\
        CC + I21K (Softmax only)        & 42.5 & 27.1 & 83.4 & 44.2 & 34.2 & 39.3 & 20.8 & 24.7 & 82.4 \\
        CC + I21K (NCE+Softmax)         & \bf 49.0 & 49.4 & 84.6 & 60.0 & 38.6 & 44.6 & 22.5 & 28.5 & \bf 82.9 \\
        CC + I21K (NCE+Softmax) + LAION & 46.9 & \bf 49.7 & \bf 85.2 & \bf 62.8 & \bf 51.5 & \bf 66.6 & \bf 25.7 & \bf 33.4 & 82.6 \\
    \bottomrule
    \end{tabular}
    }
    \end{center}
    \caption{
        {\bf Comparison of datasets \& objectives with AGD on IMP-S}.
        We integrate Conceptual Captions (CC) with contrastive (NCE) loss, and ImageNet21K (I21K) with NCE and softmax loss.
        The addition of both NCE and Softmax objectives from classification-based pretraining datasets are mutually beneficial with the retrieval-based pretraining datasets, observing best performance with the combination of both objectives.
        Further optimality is provided by adding larger, more diverse dataset like LAION-400M.
        However, we find that LAION causes optimization on contrastive objectives to become unstable, so softmax loss can greatly stabilize this noisier dataset.
    }
    \label{table:datasets}
\end{table}

%%%%%%%%%%%%%%%%%%%%%%%%%%%%%%%%%%%%%%%%%%%%%%%%%%%%%%%%%%%%
\section{Framework Modules}
\label{sec:framework-modules}

\begin{figure}[htbp]
    \centering
    \includegraphics[width=\linewidth]{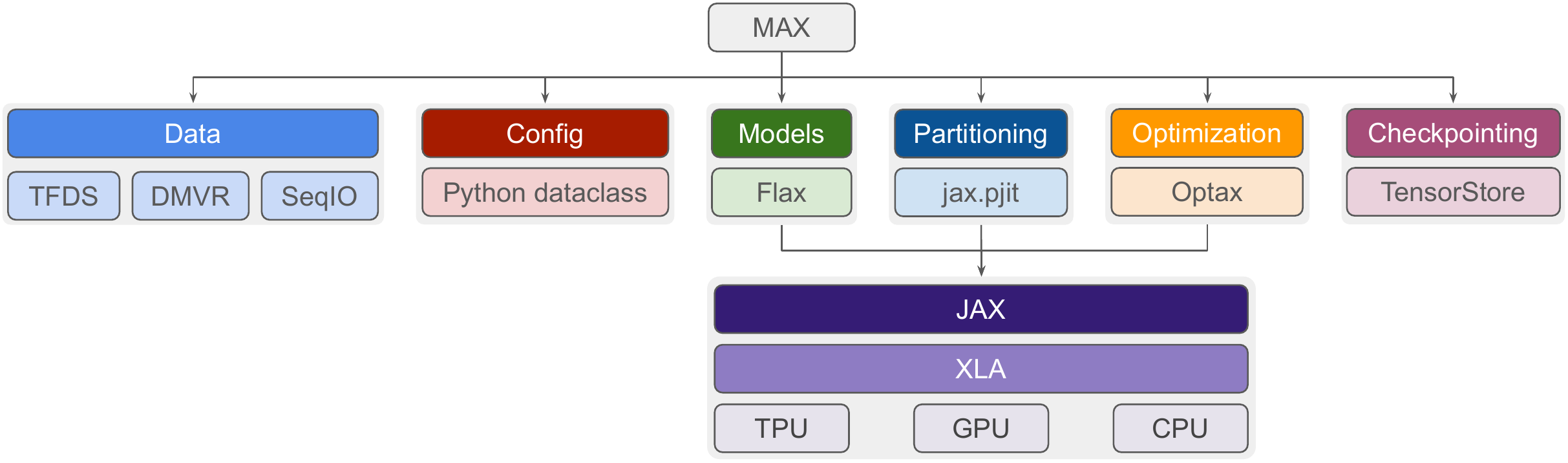}
    \caption{{\bf Modules used to implement our MAX framework.}}
    \label{figure:max-modules}
\end{figure}

To make IMP possible, we have developed a framework for AGD which we call MAX, abbreviated from \textbf{M}ulti-task Multi-modal training based on J\textbf{AX}.
MAX provides an end-to-end framework for running arbitrary multimodal data on models efficiently.
An overview of modules used in MAX is provided in Figure~\ref{figure:max-modules}.

% We begin the description of our approach through a high-level design of the framework for IMP, which we call MAX.
% Inspired by t5x~\citep{roberts2022scaling}, MAX is a multimodal framework written in JAX, built with efficient multimodal multi-task optimization in mind.
% Below, we outline some of the key features that contribute to the success of our models.

The data pipeline defines data using TensorFlow Datasets (TFDS)~\citep{tensorflow2015-whitepaper} and SeqIO~\citep{roberts2022scaling} registries for vision and language tasks.
Preprocessing of text is provided by SeqIO, while image, video, and audio preprocessors are provided by DeepMind Video Readers (DMVR)\footnote{\url{https://github.com/deepmind/dmvr}}.
Datasets are emitted from a {\tt tf.data.Dataset} object provide a key-value signature that can be tightly integrated with models.
This signature should be consistent with the model's expected input structure.
For IMP, we define named keys for each modality and emit the applicable modalities from each dataset.
Each modality can further provide optional metadata, information that specify how to properly execute or route the input to various modules.

Models are built as native Flax\footnote{\url{https://github.com/google/flax}} modules, partitioned {\tt jax.pjit}\footnote{\url{https://jax.readthedocs.io/en/latest/jax.experimental.pjit.html}}, and optimized by transforms defined in Optax\footnote{\url{https://github.com/deepmind/optax}}.
The JAX~\citep{jax2018github} framework provides a core selection of primitives that inferface with XLA\footnote{\url{https://www.tensorflow.org/xla}}, a library that compiles and optimizes computation graphs across different devices.
We use TensorStore\footnote{\url{https://github.com/google/tensorstore}} to efficiently checkpoint and restore partitioned model parameters using async parallel dispatch.
Configuration is specified according to Python dataclasses which can be overridden.
This allows the creation of many variants of datasets, models, and experiments without excessive code duplication.

On each training step, the training loop samples a dataset-objective pair, passing inputs from the dataset directly into the model.
Note that the routing of inputs across different model components is specifically avoided in the training loop logic to prevent the training process from being tied to a specific way to handle different input types.
Instead, the model itself handles the interpretation of any combination inputs provided from the dataset and produces a named collection of outputs.
Loss functions are applied in the training loop after sampling a dataset-objective pair and executing the model's forward pass.
Together, this provides a modular way to interchange datasets, models, and loss functions.

We leverage training and inference step partitioning from {\tt jax.pjit}, with further model and data parallelism abstractions provided by the t5x framework~\citep{roberts2022scaling} to partition model weights and activations across devices.
On a high level, PJIT enables the use of dynamic graph compilation at runtime across many distributed devices.
For each unique dataset-objective pair, PJIT will compile a new computation graph.
These graphs are all cached so on subsequent iterations re-compilation overhead is minimized.
This is used in conjunction with MoE to efficiently dispatch sparse weights across multiple devices while minimizing communication overhead.
In conjunction with the partitioner, we initialize states by defining a set of specs of shapes that the model should accept as input, using {\tt jax.eval\_shape}.
To efficiently run each training step, we pre-initialize the PRNG states of all training steps before any training takes place.

%%%%%%%%%%%%%%%%%%%%%%%%%%%%%%%%%%%%%%%%%%%%%%%%%%%%%%%%%%%%
\section{IMP Model Card}
\label{sec:appdx-model-card}

We present the IMP model card in Table~\ref{table:model-card}, following \cite{mitchell2019model}.

\begin{longtable}[c]{ p{.20\textwidth} | p{.62\textwidth} } 
\toprule
\multicolumn{2}{c}{\textbf{Model Summary}}      \\ \toprule
\multicolumn{1}{l|}{Model Architecture} & IMP is a multimodal sequence-to-sequence Transformer~\citep{vaswani2017attention} encoder. It takes image, video, audio and text as inputs to the encoder and produces their feature embeddings as outputs.  \\ \midrule
\multicolumn{1}{l|}{Input(s)} & RGB image, RGB video frame, audio waveform, audio spectrogram, text. %Example input: “Tracy used a piece of wire 4 feet long to support tomato plants in the garden. The wire was cut into pieces 6 inches long. How many pieces did she obtain?” 
\\ \midrule
\multicolumn{1}{l|}{Output(s)} & Feature embeddings corresponding to the inputs. %Example output (corresponding to the input shown above): “The wire was 4 feet long. This means it was 4 * 12 = 48 inches long. It was cut into pieces 6 inches long. This means she obtained 48 /6 = 8 pieces. The answer is 8.” 
\\
\toprule

\multicolumn{2}{c}{\textbf{Usage}}      \\ \toprule
\multicolumn{1}{l|}{Application} & The model is for research prototype and the current version is not available for the broader public usage.\\ 
\midrule
\multicolumn{1}{l|}{Known Caveats} & No.    \\
\toprule

\multicolumn{2}{c}{\textbf{System Type}}      \\ \toprule
\multicolumn{1}{l|}{System Description} & This is a standalone model.  \\ \midrule
\multicolumn{1}{l|}{Upstream Dependencies} & No.  \\ \midrule
\multicolumn{1}{l|}{Downstream Dependencies} & No. \\
\toprule

\multicolumn{2}{c}{\textbf{Implementation Frameworks}}      \\ \toprule
\multicolumn{1}{l|}{Hardware \& Software} & Hardware: TPU~\citep{jouppi2020domain}.    \vspace{0.1in}

Software: T5X~\citep{roberts2022scaling}, JAX~\citep{jax2018github}, Flaxformer\footnote{\url{https://github.com/google/flaxformer}}, MAX    \vspace{0.1in}

Details are in Section~\ref{sec:framework-modules}.
\\ \midrule
\multicolumn{1}{l|}{Compute Requirements} & Reported in Section~\ref{sec:ablation}. \\
\toprule

\multicolumn{2}{c}{\textbf{Model Characteristics}}      \\ \toprule
\multicolumn{1}{l|}{Model Initialization} & The model is trained from scratch with random initialization.  \\ \midrule
\multicolumn{1}{l|}{Model Status} & This is a static model trained on offline datasets.  \\ \midrule
\multicolumn{1}{l|}{Model Stats} & The largest IMP model has 2B parameters for its sparse variant and 300M parameters for its dense variant. \\
\toprule

\multicolumn{2}{c}{\textbf{Data Overview}}      \\ \toprule
\multicolumn{1}{l|}{Training dataset} &
The model is pre-trained on the following mixture of datasets:  Details are in Section~\ref{sec:training-setup}. \\\toprule
\multicolumn{1}{l|}{Evaluation and Fine-tuning Dataset} &
\begin{itemize}
\item \textbf{Image classification}: 
    CIFAR, ImageNet
\item \textbf{Video classification}: 
    UCF101, HMDB51, Kinetics400, Kinetics600, Kinetics700
\item \textbf{Audio classification}: 
    ESC
\item \textbf{Image to text / text to image retrieval}: 
    Flickr30k, COCO
\end{itemize}    \\
\toprule

\multicolumn{2}{c}{\textbf{Evaluation Results}}      \\ \toprule
\multicolumn{1}{l|}{Evaluation Results} & Reported in Section~\ref{sec:experiments}. \\ 
\toprule

\multicolumn{2}{c}{\textbf{Model Usage \& Limitations}}      \\ \toprule
\multicolumn{1}{l|}{Sensitive Use} & Reported in Section~\ref{sec:limitations} \\ \midrule
\multicolumn{1}{l|}{Known Limitations} & Reported in Section~\ref{sec:limitations}. \\ \midrule
\multicolumn{1}{l|}{Ethical Considerations \& Risks} & Reported in Section~\ref{sec:limitations}. \\

\bottomrule
\caption{IMP model card.}
\label{table:model-card}
\end{longtable}

%%%%%%%%%%%%%%%%%%%%%%%%%%%%%%%%%%%%%%%%%%%%%%%%%%%%%%%%%%%%
\section{Limitations \& Future Work}
\label{sec:limitations}

Our approach provides a promising new scaling direction that avoids many of the pitfalls when dealing with multimodal training.
However, there are still some remaining obstacles from fully realizing this approach.

We note that our model provides exceptional performance in zero-shot video understanding, but falls slightly short in zero-shot image and audio understanding.
We believe our training signals are have favored video understanding due to a combination of factors, including high incidence of vision data, a large sampling rate on vision tasks, and optimization losses converging faster on video.
With a larger set of more diverse data and tasks (e.g., text-only and audio-only pretraining), we believe we can provide further improvements on these modalities without introducing any signficant training cost.
% (see Appendix~\ref{sec:additional-results}).
%\todo{plot in appendix}

One unsolved question is how to best combine objectives during training.
We have only tested configurations of tasks that are sampled equally across training.
Instead, we can provide a more sophisticated curriculum to the model by scheduling tasks depending on the current step.
There has been work showing further efficiency and accuracy improvements when scheduling different types of tasks at various stages~\citep{wu2020multigrid, piergiovanni2023dynamic}.
% When faced with multiple modalities and objectives, this optimization space becomes extremely large, so it can be difficult to find the optimal task mixture and schedule.

Another obstacle is the use of multimodal MoE in the generative setting.
Experts-choose routing has been integral to allowing a high performance single tower encoder model, but due to its requirement to aggregate tokens across the sequence, it is not by itself suitable for causal objectives like autoregressive sequence prediction as used in language modeling.
Some additional modifications may make this possible, however.
% With some additional modifications, it may be possible to apply experts-choose routing in an autoregressive setting, allowing an encoder to also function as a decoder and train on these configurations jointly using AGD.

\clearpage

%% file: main.bbl
\begin{thebibliography}{57}
\providecommand{\natexlab}[1]{#1}
\providecommand{\url}[1]{\texttt{#1}}
\expandafter\ifx\csname urlstyle\endcsname\relax
  \providecommand{\doi}[1]{doi: #1}\else
  \providecommand{\doi}{doi: \begingroup \urlstyle{rm}\Url}\fi

\bibitem[Abadi et~al.(2015)Abadi, Agarwal, Barham, Brevdo, Chen, Citro,
  Corrado, Davis, Dean, Devin, Ghemawat, Goodfellow, Harp, Irving, Isard, Jia,
  Jozefowicz, Kaiser, Kudlur, Levenberg, Man\'{e}, Monga, Moore, Murray, Olah,
  Schuster, Shlens, Steiner, Sutskever, Talwar, Tucker, Vanhoucke, Vasudevan,
  Vi\'{e}gas, Vinyals, Warden, Wattenberg, Wicke, Yu, and
  Zheng]{tensorflow2015-whitepaper}
Mart\'{i}n Abadi, Ashish Agarwal, Paul Barham, Eugene Brevdo, Zhifeng Chen,
  Craig Citro, Greg~S. Corrado, Andy Davis, Jeffrey Dean, Matthieu Devin,
  Sanjay Ghemawat, Ian Goodfellow, Andrew Harp, Geoffrey Irving, Michael Isard,
  Yangqing Jia, Rafal Jozefowicz, Lukasz Kaiser, Manjunath Kudlur, Josh
  Levenberg, Dandelion Man\'{e}, Rajat Monga, Sherry Moore, Derek Murray, Chris
  Olah, Mike Schuster, Jonathon Shlens, Benoit Steiner, Ilya Sutskever, Kunal
  Talwar, Paul Tucker, Vincent Vanhoucke, Vijay Vasudevan, Fernanda Vi\'{e}gas,
  Oriol Vinyals, Pete Warden, Martin Wattenberg, Martin Wicke, Yuan Yu, and
  Xiaoqiang Zheng.
\newblock {TensorFlow}: Large-scale machine learning on heterogeneous systems,
  2015.
\newblock URL \url{https://www.tensorflow.org/}.
\newblock Software available from tensorflow.org.

\bibitem[Akbari et~al.(2021)Akbari, Yuan, Qian, Chuang, Chang, Cui, and
  Gong]{akbari2021vatt}
Hassan Akbari, Liangzhe Yuan, Rui Qian, Wei-Hong Chuang, Shih-Fu Chang, Yin
  Cui, and Boqing Gong.
\newblock Vatt: Transformers for multimodal self-supervised learning from raw
  video, audio and text.
\newblock \emph{NeurIPS}, 2021.

\bibitem[Baevski et~al.(2020)Baevski, Zhou, Mohamed, and
  Auli]{baevski2020wav2vec}
Alexei Baevski, Yuhao Zhou, Abdelrahman Mohamed, and Michael Auli.
\newblock wav2vec 2.0: A framework for self-supervised learning of speech
  representations.
\newblock \emph{NeurIPS}, 2020.

\bibitem[Bradbury et~al.(2018)Bradbury, Frostig, Hawkins, Johnson, Leary,
  Maclaurin, Necula, Paszke, Vander{P}las, Wanderman-{M}ilne, and
  Zhang]{jax2018github}
James Bradbury, Roy Frostig, Peter Hawkins, Matthew~James Johnson, Chris Leary,
  Dougal Maclaurin, George Necula, Adam Paszke, Jake Vander{P}las, Skye
  Wanderman-{M}ilne, and Qiao Zhang.
\newblock {JAX}: composable transformations of {P}ython+{N}um{P}y programs.
\newblock 2018.
\newblock URL \url{http://github.com/google/jax}.

\bibitem[Brown et~al.(2020)Brown, Mann, Ryder, Subbiah, Kaplan, Dhariwal,
  Neelakantan, Shyam, Sastry, Askell, et~al.]{brown2020language}
Tom Brown, Benjamin Mann, Nick Ryder, Melanie Subbiah, Jared~D Kaplan, Prafulla
  Dhariwal, Arvind Neelakantan, Pranav Shyam, Girish Sastry, Amanda Askell,
  et~al.
\newblock Language models are few-shot learners.
\newblock \emph{NeurIPS}, 2020.

\bibitem[Calvert(2001)]{calvert2001crossmodal}
Gemma~A Calvert.
\newblock Crossmodal processing in the human brain: insights from functional
  neuroimaging studies.
\newblock \emph{Cerebral cortex}, 2001.

\bibitem[Carreira et~al.(2018)Carreira, Noland, Banki-Horvath, Hillier, and
  Zisserman]{carreira2018short}
Joao Carreira, Eric Noland, Andras Banki-Horvath, Chloe Hillier, and Andrew
  Zisserman.
\newblock A short note about kinetics-600.
\newblock \emph{arXiv preprint arXiv:1808.01340}, 2018.

\bibitem[Carreira et~al.(2019)Carreira, Noland, Hillier, and
  Zisserman]{carreira2019short}
Joao Carreira, Eric Noland, Chloe Hillier, and Andrew Zisserman.
\newblock A short note on the kinetics-700 human action dataset.
\newblock \emph{arXiv preprint arXiv:1907.06987}, 2019.

\bibitem[Changpinyo et~al.(2021)Changpinyo, Sharma, Ding, and
  Soricut]{changpinyo2021conceptual}
Soravit Changpinyo, Piyush Sharma, Nan Ding, and Radu Soricut.
\newblock Conceptual 12m: Pushing web-scale image-text pre-training to
  recognize long-tail visual concepts.
\newblock In \emph{CVPR}, 2021.

\bibitem[Chen et~al.(2022)Chen, Wang, Changpinyo, Piergiovanni, Padlewski,
  Salz, Goodman, Grycner, Mustafa, Beyer, et~al.]{chen2022pali}
Xi~Chen, Xiao Wang, Soravit Changpinyo, AJ~Piergiovanni, Piotr Padlewski,
  Daniel Salz, Sebastian Goodman, Adam Grycner, Basil Mustafa, Lucas Beyer,
  et~al.
\newblock Pali: A jointly-scaled multilingual language-image model.
\newblock \emph{arXiv preprint arXiv:2209.06794}, 2022.

\bibitem[Chowdhery et~al.(2022)Chowdhery, Narang, Devlin, Bosma, Mishra,
  Roberts, Barham, Chung, Sutton, Gehrmann, et~al.]{chowdhery2022palm}
Aakanksha Chowdhery, Sharan Narang, Jacob Devlin, Maarten Bosma, Gaurav Mishra,
  Adam Roberts, Paul Barham, Hyung~Won Chung, Charles Sutton, Sebastian
  Gehrmann, et~al.
\newblock Palm: Scaling language modeling with pathways.
\newblock \emph{arXiv preprint arXiv:2204.02311}, 2022.

\bibitem[Dehghani et~al.(2023)Dehghani, Djolonga, Mustafa, Padlewski, Heek,
  Gilmer, Steiner, Caron, Geirhos, Alabdulmohsin, et~al.]{dehghani2023scaling}
Mostafa Dehghani, Josip Djolonga, Basil Mustafa, Piotr Padlewski, Jonathan
  Heek, Justin Gilmer, Andreas Steiner, Mathilde Caron, Robert Geirhos, Ibrahim
  Alabdulmohsin, et~al.
\newblock Scaling vision transformers to 22 billion parameters.
\newblock \emph{arXiv preprint arXiv:2302.05442}, 2023.

\bibitem[Dosovitskiy et~al.(2020)Dosovitskiy, Beyer, Kolesnikov, Weissenborn,
  Zhai, Unterthiner, Dehghani, Minderer, Heigold, Gelly,
  et~al.]{dosovitskiy2020image}
Alexey Dosovitskiy, Lucas Beyer, Alexander Kolesnikov, Dirk Weissenborn,
  Xiaohua Zhai, Thomas Unterthiner, Mostafa Dehghani, Matthias Minderer, Georg
  Heigold, Sylvain Gelly, et~al.
\newblock An image is worth 16x16 words: Transformers for image recognition at
  scale.
\newblock \emph{arXiv preprint arXiv:2010.11929}, 2020.

\bibitem[Driver and Noesselt(2008)]{driver2008multisensory}
Jon Driver and Toemme Noesselt.
\newblock Multisensory interplay reveals crossmodal influences on
  ‘sensory-specific’brain regions, neural responses, and judgments.
\newblock \emph{Neuron}, 2008.

\bibitem[Gemmeke et~al.(2017)Gemmeke, Ellis, Freedman, Jansen, Lawrence, Moore,
  Plakal, and Ritter]{gemmeke2017audio}
Jort~F Gemmeke, Daniel~PW Ellis, Dylan Freedman, Aren Jansen, Wade Lawrence,
  R~Channing Moore, Manoj Plakal, and Marvin Ritter.
\newblock Audio set: An ontology and human-labeled dataset for audio events.
\newblock In \emph{ICASSP}, 2017.

\bibitem[Gowda et~al.(2021)Gowda, Rohrbach, and Sevilla-Lara]{gowda2021smart}
Shreyank~N Gowda, Marcus Rohrbach, and Laura Sevilla-Lara.
\newblock Smart frame selection for action recognition.
\newblock In \emph{AAAI}, 2021.

\bibitem[Huang et~al.(2022)Huang, Xu, Li, Baevski, Auli, Galuba, Metze, and
  Feichtenhofer]{huang2022masked}
Po-Yao Huang, Hu~Xu, Juncheng Li, Alexei Baevski, Michael Auli, Wojciech
  Galuba, Florian Metze, and Christoph Feichtenhofer.
\newblock Masked autoencoders that listen.
\newblock \emph{NeurIPS}, 2022.

\bibitem[Jain et~al.(2017)Jain, Kar, et~al.]{jain2017non}
Prateek Jain, Purushottam Kar, et~al.
\newblock Non-convex optimization for machine learning.
\newblock \emph{Foundations and Trends{\textregistered} in Machine Learning},
  2017.

\bibitem[Jia et~al.(2021)Jia, Yang, Xia, Chen, Parekh, Pham, Le, Sung, Li, and
  Duerig]{jia2021scaling}
Chao Jia, Yinfei Yang, Ye~Xia, Yi-Ting Chen, Zarana Parekh, Hieu Pham, Quoc Le,
  Yun-Hsuan Sung, Zhen Li, and Tom Duerig.
\newblock Scaling up visual and vision-language representation learning with
  noisy text supervision.
\newblock In \emph{ICML}. PMLR, 2021.

\bibitem[Jouppi et~al.(2020)Jouppi, Yoon, Kurian, Li, Patil, Laudon, Young, and
  Patterson]{jouppi2020domain}
Norman~P Jouppi, Doe~Hyun Yoon, George Kurian, Sheng Li, Nishant Patil, James
  Laudon, Cliff Young, and David Patterson.
\newblock A domain-specific supercomputer for training deep neural networks.
\newblock \emph{Communications of the ACM}, 2020.

\bibitem[Kay et~al.(2017)Kay, Carreira, Simonyan, Zhang, Hillier,
  Vijayanarasimhan, Viola, Green, Back, Natsev, et~al.]{kay2017kinetics}
Will Kay, Joao Carreira, Karen Simonyan, Brian Zhang, Chloe Hillier, Sudheendra
  Vijayanarasimhan, Fabio Viola, Tim Green, Trevor Back, Paul Natsev, et~al.
\newblock The kinetics human action video dataset.
\newblock \emph{arXiv preprint arXiv:1705.06950}, 2017.

\bibitem[Kuehne et~al.(2011)Kuehne, Jhuang, Garrote, Poggio, and
  Serre]{kuehne2011hmdb}
Hildegard Kuehne, Hueihan Jhuang, Est{\'\i}baliz Garrote, Tomaso Poggio, and
  Thomas Serre.
\newblock Hmdb: a large video database for human motion recognition.
\newblock In \emph{ICCV}, 2011.

\bibitem[Likhosherstov et~al.(2021)Likhosherstov, Arnab, Choromanski, Lucic,
  Tay, Weller, and Dehghani]{likhosherstov2021polyvit}
Valerii Likhosherstov, Anurag Arnab, Krzysztof Choromanski, Mario Lucic,
  Yi~Tay, Adrian Weller, and Mostafa Dehghani.
\newblock Polyvit: Co-training vision transformers on images, videos and audio.
\newblock \emph{arXiv preprint arXiv:2111.12993}, 2021.

\bibitem[Lin et~al.(2014)Lin, Maire, Belongie, Hays, Perona, Ramanan,
  Doll{\'a}r, and Zitnick]{lin2014microsoft}
Tsung-Yi Lin, Michael Maire, Serge Belongie, James Hays, Pietro Perona, Deva
  Ramanan, Piotr Doll{\'a}r, and C~Lawrence Zitnick.
\newblock Microsoft coco: Common objects in context.
\newblock In \emph{ECCV}, 2014.

\bibitem[Miech et~al.(2019)Miech, Zhukov, Alayrac, Tapaswi, Laptev, and
  Sivic]{miech2019howto100m}
Antoine Miech, Dimitri Zhukov, Jean-Baptiste Alayrac, Makarand Tapaswi, Ivan
  Laptev, and Josef Sivic.
\newblock Howto100m: Learning a text-video embedding by watching hundred
  million narrated video clips.
\newblock In \emph{ICCV}, 2019.

\bibitem[Mindermann et~al.(2022)Mindermann, Brauner, Razzak, Sharma, Kirsch,
  Xu, H{\"o}ltgen, Gomez, Morisot, Farquhar, et~al.]{mindermann2022prioritized}
S{\"o}ren Mindermann, Jan~M Brauner, Muhammed~T Razzak, Mrinank Sharma, Andreas
  Kirsch, Winnie Xu, Benedikt H{\"o}ltgen, Aidan~N Gomez, Adrien Morisot,
  Sebastian Farquhar, et~al.
\newblock Prioritized training on points that are learnable, worth learning,
  and not yet learnt.
\newblock In \emph{ICML}. PMLR, 2022.

\bibitem[Mitchell et~al.(2019)Mitchell, Wu, Zaldivar, Barnes, Vasserman,
  Hutchinson, Spitzer, Raji, and Gebru]{mitchell2019model}
Margaret Mitchell, Simone Wu, Andrew Zaldivar, Parker Barnes, Lucy Vasserman,
  Ben Hutchinson, Elena Spitzer, Inioluwa~Deborah Raji, and Timnit Gebru.
\newblock Model cards for model reporting.
\newblock In \emph{Proceedings of the conference on fairness, accountability,
  and transparency}, 2019.

\bibitem[Mustafa et~al.(2022)Mustafa, Riquelme, Puigcerver, Jenatton, and
  Houlsby]{mustafa2022multimodal}
Basil Mustafa, Carlos Riquelme, Joan Puigcerver, Rodolphe Jenatton, and Neil
  Houlsby.
\newblock Multimodal contrastive learning with limoe: the language-image
  mixture of experts.
\newblock In \emph{NeurIPS}, 2022.

\bibitem[Nagrani et~al.(2022)Nagrani, Seo, Seybold, Hauth, Manen, Sun, and
  Schmid]{nagrani2022learning}
Arsha Nagrani, Paul~Hongsuck Seo, Bryan Seybold, Anja Hauth, Santiago Manen,
  Chen Sun, and Cordelia Schmid.
\newblock Learning audio-video modalities from image captions.
\newblock \emph{arXiv preprint arXiv:2204.00679}, 2022.

\bibitem[Ni et~al.(2022)Ni, Peng, Chen, Zhang, Meng, Fu, Xiang, and
  Ling]{ni2022expanding}
Bolin Ni, Houwen Peng, Minghao Chen, Songyang Zhang, Gaofeng Meng, Jianlong Fu,
  Shiming Xiang, and Haibin Ling.
\newblock Expanding language-image pretrained models for general video
  recognition.
\newblock In \emph{ECCV}, 2022.

\bibitem[Pascal et~al.(2021)Pascal, Michiardi, Bost, Huet, and
  Zuluaga]{pascal2021improved}
Lucas Pascal, Pietro Michiardi, Xavier Bost, Benoit Huet, and Maria~A Zuluaga.
\newblock Improved optimization strategies for deep multi-task networks.
\newblock \emph{arXiv preprint arXiv:2109.11678}, 2021.

\bibitem[Pham et~al.(2021)Pham, Dai, Ghiasi, Kawaguchi, Liu, Yu, Yu, Chen,
  Luong, Wu, et~al.]{pham2021combined}
Hieu Pham, Zihang Dai, Golnaz Ghiasi, Kenji Kawaguchi, Hanxiao Liu, Adams~Wei
  Yu, Jiahui Yu, Yi-Ting Chen, Minh-Thang Luong, Yonghui Wu, et~al.
\newblock Combined scaling for open-vocabulary image classification.
\newblock \emph{arXiv preprint arXiv: 2111.10050}, 2021.

\bibitem[Piczak(2015)]{piczak2015esc}
Karol~J Piczak.
\newblock Esc: Dataset for environmental sound classification.
\newblock In \emph{ACM MM}, 2015.

\bibitem[Piergiovanni et~al.(2023)Piergiovanni, Kuo, Li, and
  Angelova]{piergiovanni2023dynamic}
AJ~Piergiovanni, Weicheng Kuo, Wei Li, and Anelia Angelova.
\newblock Dynamic pretraining of vision-language models, 2023.
\newblock URL \url{https://openreview.net/forum?id=QcffIcjq8bl}.

\bibitem[Radford et~al.(2021)Radford, Kim, Hallacy, Ramesh, Goh, Agarwal,
  Sastry, Askell, Mishkin, Clark, et~al.]{radford2021learning}
Alec Radford, Jong~Wook Kim, Chris Hallacy, Aditya Ramesh, Gabriel Goh,
  Sandhini Agarwal, Girish Sastry, Amanda Askell, Pamela Mishkin, Jack Clark,
  et~al.
\newblock Learning transferable visual models from natural language
  supervision.
\newblock In \emph{ICML}. PMLR, 2021.

\bibitem[Raffel et~al.(2020)Raffel, Shazeer, Roberts, Lee, Narang, Matena,
  Zhou, Li, Liu, et~al.]{raffel2020exploring}
Colin Raffel, Noam Shazeer, Adam Roberts, Katherine Lee, Sharan Narang, Michael
  Matena, Yanqi Zhou, Wei Li, Peter~J Liu, et~al.
\newblock Exploring the limits of transfer learning with a unified text-to-text
  transformer.
\newblock \emph{JMLR}, 2020.

\bibitem[Ridnik et~al.(2021)Ridnik, Ben-Baruch, Noy, and
  Zelnik-Manor]{ridnik2021imagenet}
Tal Ridnik, Emanuel Ben-Baruch, Asaf Noy, and Lihi Zelnik-Manor.
\newblock Imagenet-21k pretraining for the masses.
\newblock \emph{arXiv preprint arXiv:2104.10972}, 2021.

\bibitem[Riquelme et~al.(2021)Riquelme, Puigcerver, Mustafa, Neumann, Jenatton,
  Susano~Pinto, Keysers, and Houlsby]{riquelme2021scaling}
Carlos Riquelme, Joan Puigcerver, Basil Mustafa, Maxim Neumann, Rodolphe
  Jenatton, Andr{\'e} Susano~Pinto, Daniel Keysers, and Neil Houlsby.
\newblock Scaling vision with sparse mixture of experts.
\newblock \emph{NeurIPS}, 2021.

\bibitem[Roberts et~al.(2022)Roberts, Chung, Levskaya, Mishra, Bradbury, Andor,
  Narang, Lester, Gaffney, Mohiuddin, et~al.]{roberts2022scaling}
Adam Roberts, Hyung~Won Chung, Anselm Levskaya, Gaurav Mishra, James Bradbury,
  Daniel Andor, Sharan Narang, Brian Lester, Colin Gaffney, Afroz Mohiuddin,
  et~al.
\newblock Scaling up models and data with t5x and seqio.
\newblock \emph{arXiv preprint arXiv:2203.17189}, 2022.

\bibitem[Russakovsky et~al.(2015)Russakovsky, Deng, Su, Krause, Satheesh, Ma,
  Huang, Karpathy, Khosla, Bernstein, Berg, and Fei-Fei]{ILSVRC15}
Olga Russakovsky, Jia Deng, Hao Su, Jonathan Krause, Sanjeev Satheesh, Sean Ma,
  Zhiheng Huang, Andrej Karpathy, Aditya Khosla, Michael Bernstein,
  Alexander~C. Berg, and Li~Fei-Fei.
\newblock {ImageNet Large Scale Visual Recognition Challenge}.
\newblock \emph{IJCV}, 2015.

\bibitem[Schuhmann et~al.(2021)Schuhmann, Vencu, Beaumont, Kaczmarczyk, Mullis,
  Katta, Coombes, Jitsev, and Komatsuzaki]{schuhmann2021laion}
Christoph Schuhmann, Richard Vencu, Romain Beaumont, Robert Kaczmarczyk,
  Clayton Mullis, Aarush Katta, Theo Coombes, Jenia Jitsev, and Aran
  Komatsuzaki.
\newblock Laion-400m: Open dataset of clip-filtered 400 million image-text
  pairs.
\newblock \emph{arXiv preprint arXiv:2111.02114}, 2021.

\bibitem[Shen et~al.(2023)Shen, Yao, Li, Darrell, Keutzer, and
  He]{shen2023scaling}
Sheng Shen, Zhewei Yao, Chunyuan Li, Trevor Darrell, Kurt Keutzer, and Yuxiong
  He.
\newblock Scaling vision-language models with sparse mixture of experts.
\newblock \emph{arXiv preprint arXiv:2303.07226}, 2023.

\bibitem[Smith and Gasser(2005)]{smith2005development}
Linda Smith and Michael Gasser.
\newblock The development of embodied cognition: Six lessons from babies.
\newblock \emph{Artificial life}, 2005.

\bibitem[Soomro et~al.(2012)Soomro, Zamir, and Shah]{soomro2012ucf101}
Khurram Soomro, Amir~Roshan Zamir, and Mubarak Shah.
\newblock Ucf101: A dataset of 101 human actions classes from videos in the
  wild.
\newblock \emph{arXiv preprint arXiv:1212.0402}, 2012.

\bibitem[Srinivasan et~al.(2021)Srinivasan, Raman, Chen, Bendersky, and
  Najork]{srinivasan2021wit}
Krishna Srinivasan, Karthik Raman, Jiecao Chen, Michael Bendersky, and Marc
  Najork.
\newblock Wit: Wikipedia-based image text dataset for multimodal multilingual
  machine learning.
\newblock In \emph{SIGIR}, 2021.

\bibitem[Stroud et~al.(2020)Stroud, Lu, Sun, Deng, Sukthankar, Schmid, and
  Ross]{stroud2020learning}
Jonathan~C Stroud, Zhichao Lu, Chen Sun, Jia Deng, Rahul Sukthankar, Cordelia
  Schmid, and David~A Ross.
\newblock Learning video representations from textual web supervision.
\newblock \emph{arXiv preprint arXiv:2007.14937}, 2020.

\bibitem[Vaswani et~al.(2017)Vaswani, Shazeer, Parmar, Uszkoreit, Jones, Gomez,
  Kaiser, and Polosukhin]{vaswani2017attention}
Ashish Vaswani, Noam Shazeer, Niki Parmar, Jakob Uszkoreit, Llion Jones,
  Aidan~N Gomez, {\L}ukasz Kaiser, and Illia Polosukhin.
\newblock Attention is all you need.
\newblock \emph{NeurIPS}, 2017.

\bibitem[Wang et~al.(2022)Wang, Bao, Dong, Bjorck, Peng, Liu, Aggarwal,
  Mohammed, Singhal, Som, et~al.]{wang2022image}
Wenhui Wang, Hangbo Bao, Li~Dong, Johan Bjorck, Zhiliang Peng, Qiang Liu, Kriti
  Aggarwal, Owais~Khan Mohammed, Saksham Singhal, Subhojit Som, et~al.
\newblock Image as a foreign language: Beit pretraining for all vision and
  vision-language tasks.
\newblock \emph{arXiv preprint arXiv:2208.10442}, 2022.

\bibitem[Wu et~al.(2020)Wu, Girshick, He, Feichtenhofer, and
  Krahenbuhl]{wu2020multigrid}
Chao-Yuan Wu, Ross Girshick, Kaiming He, Christoph Feichtenhofer, and Philipp
  Krahenbuhl.
\newblock A multigrid method for efficiently training video models.
\newblock In \emph{CVPR}, 2020.

\bibitem[Wu et~al.(2022{\natexlab{a}})Wu, Liang, Han, Akbari, Wang, and
  Yu]{wu2022scaling}
Junru Wu, Yi~Liang, Feng Han, Hassan Akbari, Zhangyang Wang, and Cong Yu.
\newblock Scaling multimodal pre-training via cross-modality gradient
  harmonization.
\newblock In \emph{NeurIPS}, 2022{\natexlab{a}}.

\bibitem[Wu et~al.(2022{\natexlab{b}})Wu, Sun, and Ouyang]{wu2022transferring}
Wenhao Wu, Zhun Sun, and Wanli Ouyang.
\newblock Transferring textual knowledge for visual recognition.
\newblock \emph{arXiv preprint arXiv:2207.01297}, 2022{\natexlab{b}}.

\bibitem[Wu et~al.(2022{\natexlab{c}})Wu, Wang, Luo, Wang, Yang, and
  Ouyang]{wu2022bidirectional}
Wenhao Wu, Xiaohan Wang, Haipeng Luo, Jingdong Wang, Yi~Yang, and Wanli Ouyang.
\newblock Bidirectional cross-modal knowledge exploration for video recognition
  with pre-trained vision-language models.
\newblock \emph{arXiv preprint arXiv:2301.00182}, 2022{\natexlab{c}}.

\bibitem[Yan et~al.(2022)Yan, Zhu, Wang, Cao, Zhang, Ghosh, Wu, and
  Yu]{yan2022video}
Shen Yan, Tao Zhu, Zirui Wang, Yuan Cao, Mi~Zhang, Soham Ghosh, Yonghui Wu, and
  Jiahui Yu.
\newblock Video-text modeling with zero-shot transfer from contrastive
  captioners.
\newblock \emph{arXiv preprint arXiv:2212.04979}, 2022.

\bibitem[Young et~al.(2014)Young, Lai, Hodosh, and Hockenmaier]{young2014image}
Peter Young, Alice Lai, Micah Hodosh, and Julia Hockenmaier.
\newblock From image descriptions to visual denotations: New similarity metrics
  for semantic inference over event descriptions.
\newblock \emph{Transactions of the Association for Computational Linguistics},
  2014.

\bibitem[Yu et~al.(2022)Yu, Wang, Vasudevan, Yeung, Seyedhosseini, and
  Wu]{yu2022coca}
Jiahui Yu, Zirui Wang, Vijay Vasudevan, Legg Yeung, Mojtaba Seyedhosseini, and
  Yonghui Wu.
\newblock Coca: Contrastive captioners are image-text foundation models.
\newblock \emph{arXiv preprint arXiv:2205.01917}, 2022.

\bibitem[Zhai et~al.(2022)Zhai, Kolesnikov, Houlsby, and
  Beyer]{zhai2022scaling}
Xiaohua Zhai, Alexander Kolesnikov, Neil Houlsby, and Lucas Beyer.
\newblock Scaling vision transformers.
\newblock In \emph{CVPR}, 2022.

\bibitem[Zhou et~al.(2022)Zhou, Lei, Liu, Du, Huang, Zhao, Dai, Chen, Le, and
  Laudon]{zhou2022mixture}
Yanqi Zhou, Tao Lei, Hanxiao Liu, Nan Du, Yanping Huang, Vincent Zhao, Andrew
  Dai, Zhifeng Chen, Quoc Le, and James Laudon.
\newblock Mixture-of-experts with expert choice routing.
\newblock In \emph{NeurIPS}, 2022.

\end{thebibliography}
